\documentclass{article}

\usepackage{microtype}
\usepackage{graphicx}
\usepackage{subfigure}
\usepackage{booktabs} % for professional tables

\usepackage{hyperref}

\usepackage[accepted]{icml2023}

\usepackage{amsmath}
\usepackage{amssymb}
\usepackage{mathtools}
\usepackage{amsthm}
\usepackage{xcolor}
\usepackage{times}
\usepackage{soul}
\usepackage{url}
\usepackage[utf8]{inputenc}
\usepackage[small]{caption}
\usepackage{graphicx}
\usepackage{booktabs}
\usepackage{algorithm}
\usepackage[ruled,noend,vlined,algo2e, linesnumbered]{algorithm2e}
\usepackage{setspace}
\DeclareMathOperator*{\argmax}{arg\,max}
\DeclareMathOperator*{\argmin}{arg\,min}
\newcommand{\algoname}{K-SHAP}
\usepackage{multirow}

\usepackage[capitalize,noabbrev]{cleveref}

\theoremstyle{plain}

\theoremstyle{definition}

\theoremstyle{remark}

\usepackage[textsize=tiny]{todonotes}

\icmltitlerunning{{\algoname}: Policy Clustering Algorithm for Anonymous State-Action Pairs}

\begin{document}
\twocolumn[
\icmltitle{{\algoname}: Policy Clustering Algorithm for \\ Anonymous Multi-Agent State-Action Pairs}

\begin{icmlauthorlist}
\icmlauthor{Andrea Coletta}{xxx}
\icmlauthor{Svitlana Vyetrenko}{yyy}
\icmlauthor{Tucker Balch}{yyy}
\end{icmlauthorlist}

\icmlaffiliation{xxx}{J.P. Morgan AI Research, London, UK.}
\icmlaffiliation{yyy}{J.P. Morgan AI Research, New York, USA}

\icmlcorrespondingauthor{Andrea Coletta}{andrea.coletta@jpmchase.com}

\icmlkeywords{Machine Learning, ICML}

\vskip 0.3in
]
\printAffiliationsAndNotice{}

\begin{abstract}
% Problem and Importance
Learning agent behaviors from observational data has shown to improve our understanding of their decision-making processes, advancing our ability to explain their interactions with the environment and other agents. 
% Existing work / background
While multiple learning techniques have been proposed in the literature, there is one particular setting that has not been explored yet: multi agent systems where agent identities remain anonymous.
For instance, in financial markets labeled data that identifies market participant strategies is typically proprietary, and only the anonymous state-action pairs that
result from the interaction of multiple market participants are publicly
available. As a result,
sequences of agent actions are not observable, restricting the applicability of existing work.
% Solution
In this paper, we propose a Policy Clustering algorithm, called {\algoname}, that learns to group anonymous state-action pairs according to the agent policies.
% Insight
We frame the problem as an Imitation Learning (IL) task, and we learn a \textit{world-policy} able to mimic all the agent behaviors upon different environmental states. We leverage the \textit{world-policy} to explain each anonymous observation through an additive feature attribution method called SHAP (SHapley Additive exPlanations). Finally, by clustering the explanations we show that we are able to identify different agent policies and group observations accordingly. 
% Evidence 
We evaluate our approach on simulated synthetic market data and a real-world financial dataset. We show that our proposal significantly and consistently outperforms the existing methods, identifying different agent strategies.
\end{abstract}

\section{Introduction}\label{sec:intro}
% Intro to problem
Insect colonies, animal swarms, and human societies are examples of the complex multi-agent systems in nature. Each agent in these systems develops a behavioral strategy to take suitable actions in response to environmental changes. 
Animals learn different behaviors to survive and exploit available resources (e.g., food), while humans may develop complex behaviors to \mbox{efficiently interact and pursue their own goals.}

% why important
The study and understanding of human and animal behaviors has been a fundamental problem in both computer and behavioral sciences~\cite{cichos2020machine}. In particular, understanding agent behaviors from observational data is essential to study, predict and simulate their behaviors~\cite{li2020generative,suo2021trafficsim}.

% Current approaches 
Existing work borrows tools from the reinforcement learning literature to learn agent strategies~\cite{hussein2017imitation,ho2016generative,song2018multi,fu2021evaluating} using Markov Decision Processes (MDPs) as an efficient mathematical framework to formulate the problem.
In particular, Inverse Reinforcement Learning (IRL)~\cite{ng2000algorithms} has been widely studied to characterize the decision-making behavior of animals and humans. IRL aims at recovering a reward function that explains the agent goal. It has been used to study worm behavioral strategies~\cite{yamaguchi2018identification}; to model the behaviors of mice exploring a labyrinth~\cite{ashwooddynamic}; to identify and capture the behavior of troll accounts in social networks~\cite{luceri2020detecting}. Imitation Learning (IL)~\cite{hussein2017imitation} is another technique that has demonstrated great success in modeling agent behaviors, learning directly from their trajectories (state-action pairs). Recently, IL has shown to learn realistic driving behaviors from human demonstrations~\cite{suo2021trafficsim}. Finally, Hidden Markov Models (HMMs) have been extensively used to analyze temporal dynamics and model agent behaviors from observed sequential data. HMMs have been used to study different animal behaviors, including honey bees~\cite{feldman2004modeling} and mice~\cite{jiang2018context}.

% Challenge
While all these techniques assume that clear observation sequences for each agent are given, some critical domains with privacy concerns may provide only anonymous state-action pairs as data. 
For example, military operations are often anonymous to conceal strategies to opponents, especially in case of cyber-attacks or when illegal activities are operated (e.g., use of chemical or biological weapons)~\cite{koblentz2019chemical,rid2015attributing}. In financial markets labeled data that identifies market participants is typically proprietary, and publicly available transaction data is typically anonymous~\cite{totalview}. However, understanding market participant strategies for the purpose of ensuring that markets are orderly and compliant with regulations is necessary~\cite{hagstromer2013diversity,kirilenko2017flash,Wang2021spoofing}. 
For instance, in \cite{kirilenko2017flash} the authors were able to study and demonstrate that high frequency traders did not contribute to the 2010 flash crash using audit trail transaction-level data, which reveals the identities of market participants.
Nevertheless, it is most common that only the anonymous state-action pairs without any agent identifiers are publicly available from exchanges~\cite{totalview}. 

It is well known that thousands of individual market participants can be broadly assigned to a small number of distinct behavioral strategies~\cite{kirilenko2017flash,vyetrenko2020get}. For example, multiple market agents might trade on momentum signals even though these momentum signals are of different magnitudes. 
Similarly, all market makers are required to place both buy and sell orders to provide liquidity in the markets, even though individual market makers might act upon different proprietary signals to do that.
Previous work has used IRL with reward clustering to distinguish high frequency from other trading strategies in simulated (but not the real) markets~\cite{yang2012behavior}. However, the proposed method requires the inventory level and the labeled sequence of actions for each trader, which are usually unknown in real markets. We are not aware of any other work that would allow us to identify the individual agents or their strategies from anonymous state-action pairs. 

\begin{figure}[t]
\includegraphics[trim={0 0 0 0 }, width=1.05\linewidth]{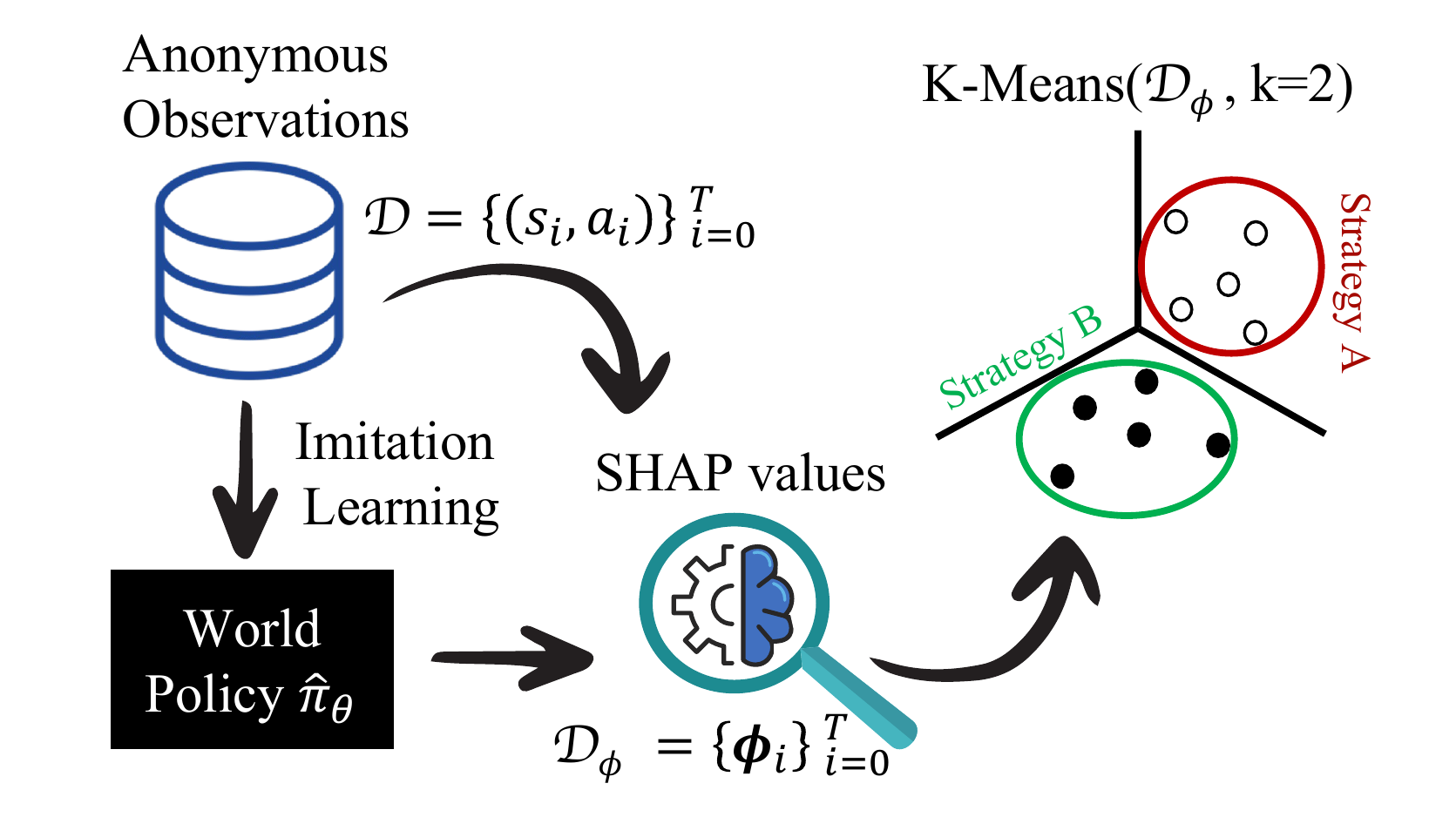}
\centering
\caption{The three major phases of {\algoname}.}\label{fig:framework}
% \vspace{-0.1in}
\end{figure}

% addressed problem
To address the challenge of learning agent strategies from anonymous state-action pairs, in this paper we propose a novel policy-clustering method -- in which we group state-action observations that belong to agents sharing the same behavior or policy. These clusters describe the different behaviors of agents, enabling further studies and analysis.
\\
In detail, we propose {\algoname}, a Policy Clustering algorithm for anonymous state-action pairs. {\algoname} comprises three major phases shown in Figure \ref{fig:framework}. First, we model the problem as an Imitation Learning task to learn a \textit{world-policy}~\cite{coletta2022learning} using supervised learning over the anonymous state-action pairs. The \textit{world-policy} is an extension of the \textit{world-model} approach~\cite{ha2018recurrent} to state-action pairs: it emulates collective agent behaviors, learning their actions (output) in response to the different environmental states (input). Then, we interpret the world-policy to explain each anonymous observation through an additive feature attribution method, namely SHAP (SHapley Additive exPlanations)~\cite{lundberg2017unified}. Finally, we show that the explanations, called SHAP values, better reveal the intrinsic clustering structure among data; and by applying a K-Means algorithm~\cite{hartigan1979algorithm} in the new SHAP values space, we can group the anonymous observations into $K$ clusters, which reflect the real agent strategies. 

To support the soundness of our method, we experimentally evaluate it with market data.  
We first consider synthetic data from a multi-agent simulator labeled with agent IDs to serve as a ground truth~\cite{byrd2019abides,liu2022biased}, then we consider real anonymous market data from NASDAQ stock exchange~\cite{totalview}. For real data we evaluate the 
\textit{Utility} and \textit{Silhouette}, as ground truth labels are not available. In particular, by \textit{Utility} we mean the ability to improve a specific downstream task (i.e., supervised learning) when trained on the clusters rather than original demonstrations.
We compare our proposal against state-of-art clustering algorithms, including Deep Clustering Network (DCN)~\cite{mukherjee2019clustergan} and ClusterGAN~\cite{mukherjee2019clustergan}. We also adapt $\Sigma$-GIRL~\cite{ramponi2020truly}, a  Multiple-Intent IRL clustering algorithm, to anonymous state-action pairs. We show that we significantly and consistently outperform the existing work, and that {\algoname} clusters anonymous observations according to the agent behaviors. {\algoname} outperforms existing methods by a factor of 2 on all the performance metrics.

\subsection{Main contributions}
To the best of our knowledge, this is the first work that addresses the problem of policy clustering with anonymous state-action pairs as observations from multiple agents. 
We summarize the main contributions of our paper as follows:
% \vspace{0.05in}
\begin{itemize}
\item  We formalize the k-policy clustering problem under anonymous multi-agent state-action pairs.
\item   We propose {\algoname}, a Policy Clustering Algorithm which leverages IL and SHAP values to identify k distinct behavioral clusters.
\item   We experimentally evaluate {\algoname} against existing work, showing that it achieves good performance in all performance metrics, with both synthetic and real-world data. In particular, we show that {\algoname} can group market observations according to the trader strategies, improving our understanding of market participant and their impact. 
\end{itemize}

\section{Preliminaries and Problem Formulation}

\subsection{Agent behaviors as Markov Decision Processes} We consider a multi-agent system with $N$ autonomous agents interacting in a given environment. We adopt Markov Decision Processes (MDPs) as a natural underlying decision model for each single agent. A MDP is a tuple $\mathcal{M} = (\mathcal{S}, \mathcal{A}, \mathcal{P}, \mathcal{R})$ composed by: the state space $\mathcal{S}$; the action space $\mathcal{A}$; a transition function $\mathcal{P}: \mathcal{S} \times \mathcal{A} \rightarrow \Delta(\mathcal{S})$, where $\Delta(\mathcal{S})$ denotes the probability distributions over state space $\mathcal{S}$; and a reward function $\mathcal{R}: \mathcal{S} \times \mathcal{A} \rightarrow \mathbb{R}$. 
\\
We consider that each agent follows a policy $\pi$ (i.e., behavior) to maximize the expected reward over time:
% \vspace{-0.05in}
\begin{equation}
J(\pi) = \mathbb{E}\left[\sum_{t=0}^{T-1} \gamma^t \mathcal{R}(s_t, a_t)  \right] 
% \vspace{-0.05in}
\end{equation}%% \vspace{-0.02in}
where $\gamma \in [0,1]$ is the discounted factor that can be used to prioritize early actions. We consider a general policy $\pi : S \rightarrow A$ that describes the preferred action $a \in \mathcal{A}$ in the state $s \in \mathcal{S}$. 

% \vspace{-0.1in}
\subsection{Agent observations} In our problem setting, we do not have direct access to the agent rewards or policies, we consider instead a more general case of an MDP without reward (MDP$ \setminus \{\mathcal{R}\}$). In this setting the reward $\mathcal{R}$ is unknown, and agent behaviors are represented by a set of trajectories $\mathcal{D} = \{\xi_0, \ldots, \xi_t\}$ for each agent, where a trajectory is a sequence of state-action pairs $\xi_i =\{(s_0, a_0), \ldots, (s_k, a_k)\}$. 
However, we have an additional confidentiality constraint: agents act anonymously and their observations do not contain any identifier. Thus, we are not able to identify a trajectory $\xi$ for a single agent, but observational data is anonymously gathered from all the agents into a new set $\mathcal{D} = \{(s_i, a_i)\}_{i=1}^{T}$ over a time $T$. Notice that the collected state-action pairs $(s_i,a_i)$ do not contain any information about which anonymous agent generated the action $a_i$ at state $s_i$, preventing the identification of state-action sequences generated by an agent.
This scenario is typical of military or financial domains, where agents want to hide their strategies for tactical reasons. In particular, in financial domains, the \textit{actions} are the orders submitted by traders according to the current financial market \textit{state}. 
\\
Anonymous observations exponentially increase the complexity of behavioral studies as identifying an agent behavior with existing approaches is almost impracticable: considering $k$ agents and $n$ observations, we have $O(k^n)$ possible assignment of the observations to reconstruct the agent trajectories needed by most of the existing methodologies. 

% \vspace{-0.1in}
\subsection{Reward functions to model agent behaviors} Given an MDP formulation, a classic approach to identify agent behaviors is to study their reward function $\mathcal{R}$. This function is the most succinct representation of the agent goals, which define their \textit{extrinsic motivation} to act~\cite{chentanez2004intrinsically}. Therefore, rewards help to study agent behaviors and can be used to solve \textit{intent-clustering}~\cite{ramponi2020truly,babes2011apprenticeship} %and \textit{policy clustering}~\cite{yang2012behavior} problems,
in which we aim at identifying groups of agents sharing the same goal. % and behavior, respectively. 
Considering a MDP$ \setminus \{\mathcal{R}\}$, previous work employs IRL to recover the \textit{unknown} reward function  \mbox{ $\mathcal{R}: \mathcal{S} \times \mathcal{A} \rightarrow \mathbb{R}$} from the agent demonstrations~\cite{ramponi2020truly,yamaguchi2018identification,yang2012behavior}.
In the general case, the unknown reward function $\mathcal{R}_{\boldsymbol{\omega}}(s,a)$ can be defined as a linear combination of $q$ weighted features:
$\boldsymbol{\omega}^T \boldsymbol{\phi}(s,a), \ \ \boldsymbol{\omega}   \in \mathbb{R}^q$, where $\boldsymbol{\phi}(s,a)$ is the feature function and $\boldsymbol{\omega}^T$ the feature weights.  
\\
However, in IRL the problem is ill-posed: even under perfect knowledge with demonstrations from an optimal policy, there exist many solutions (reward functions) for which a given behavior is optimal~\cite{ng2000algorithms}. In general, the real reward function is not identifiable~\cite{cao2021identifiability}. Most important, IRL requires observations as sequences of state-action pairs for each agent, while we consider domains in which state-action pairs are anonymous.

\subsection{Learning a policy to study agent behaviors}%\label{sec:shap}
Another simple, yet effective, approach to study the behavior of an agent is to directly describe their mapping from state features to actions, as the most parsimonious description of the policy itself.
Recently, IL~\cite{hussein2017imitation} has demonstrated that it is possible to reconstruct the agent policies directly from state-action pair observations. In particular, in IL we aim at learning an optimal policy $\pi_{\theta}$ for an agent $i$ according to its demonstrations:
% \vspace{-0.05in}
\begin{equation}
    \argmax_{\theta} \mathbb{E}_{s \sim d^{\pi_{i}}, a \sim \pi_i( \cdot | s)} [ log \ \pi_{\theta} (a | s) ]
    % \vspace{-0.05in}
\end{equation}
Notice that $d^{\pi_{i}}$ denotes the distribution over states induced by agent policy $\pi_i$. Once we reconstruct the policy $\pi_i$, we directly study the behavior of the agent $i$ by investigating how the policy maps state to actions. 
\\
In fact, for a simple policy the best explanation is often the model $\pi_i$ itself, as a parsimonious description of the agent behavior. However, models describing complex policies cannot be easily studied and explained. In such cases, we can borrow techniques from explainable AI to investigate the behavior of the model~\cite{dwivedi2022explainable}. 

\subsection{SHAP (SHapley Additive exPlanations)}\label{sec:shap}
A recent successful technique for explainability is the SHAP values method (SHapley Additive exPlanations)~\cite{lundberg2017unified}. This method belongs to the additive feature attribution methods, which are local methods designed to explain a single prediction $f(x)$ based on the input $x$~\cite{ribeiro2016should}. These methods approximate the real model $f$ using a simpler \textit{explanation model} $g$. The model $g$ tries to guarantee that $g(x') \approx f(h_x(x'))$, where $x'$ is a simplified input for $g$, and $h_x$ is a mapping function to reconstruct the original input $x=h_x(x')$.
The \textit{explanation model} is then a linear combination of binary variables $x'$:
% \vspace{-0.1in}
\begin{equation}
    g(x') = \phi_0 + \sum_{i=1}^{m} \phi_i \cdot x_i'
    % \vspace{-0.05in}
\end{equation}%% \vspace{-0.1in}
where $m$ is the number of simplified input features, and $\phi_i \in \mathbb{R}$ measures the contribution of each feature to the model output. The sum of all feature contributions approximates the original model output $f(x)$. In particular, SHAP values method uses classic game theory to explain the model predictions. Given a set of features $F$ the method computes the contribution $\phi_i$ of each feature $i \in F$ by evaluating the model with and without such a feature. To fairly account for the effects of the withholding of a feature among the others, SHAP computes the average contribution of a feature $i$ considering the model over all the possible subsets $S \subseteq F \setminus {i}$:
% \vspace{-0.05in}
$$
\small
    \phi_i = \sum_{S \subseteq F \setminus {i}} \frac{|S|!(|F| - |S| - 1)!}{|F|!}
    \bigg[f_{S \cup \{i\}}(x_{S \cup \{i\}}) - f_{S}(x_{S}) \bigg]
$$
where $x_S$ represents the input features in the subset $S$, and $f_{S}$ is the model trained on the subset of features $S$ or its approximation. Therefore, the SHAP values describe how the input features (state) contribute to the output (action). We denote with $\Phi \in \mathbb{R}^m$ the new space of SHAP values,  and with $\boldsymbol{\phi} \in \Phi$ the SHAP values vector for an input $x$.

While IL and SHAP values provide a promising approach to study the behavior of an agent, IL requires a trajectory $\xi$, or a set of state-action pairs, for each agent to approximate the policy correctly. 
Therefore, we cannot directly combine these two methods in our setting, as state-action pairs are anonymous without any agent identifier. In the following sections, we will show that if we are able to learn a \textit{world-policy} describing all the agent behaviors, SHAP values are naturally well suited to describe each observation as the market state contribution to the action, which characterizes the different trader strategies (see Section \ref{sec:intro}).

% \vspace{-0.05in}
\subsection{Policy-Clustering under anonymous state-action pairs}
In the general policy-clustering setting we consider a set of $n$ agents $\boldsymbol{\mathrm{A}}=\{ \mathrm{A}_0, \ldots, \mathrm{A}_n\}$  following a finite unknown set of policies $\boldsymbol{\pi} = \{\pi_0, \ldots, \pi_k\}$ such that $k < n$ (i.e., multiple agents may have the same behavior $\pi_i$). The goal of policy-clustering is to group the agents $\boldsymbol{\mathrm{A}}$ into $k$ clusters, according to their policies from the observational data  \mbox{$\mathcal{D} = \{\xi_0, \ldots, \xi_t\}_{i=0}^n$}. Agents in the same cluster will follow the same strategy or behavior.
Unlike the general formulation, we relax the assumption of knowing agent identities and trajectories. Thus, the goal of policy-clustering under anonymous state-action pairs is to group the anonymous observational data $ \mathcal{D} = \{(s_i, a_i)\}_{i=0}^T$ into $k$ clusters, such that observations in the same cluster belong to agents sharing the same policy $\pi$. 

\section{{\algoname} Framework}
This section introduces our main contribution, {\algoname}, a Policy Clustering algorithm that learns to cluster the anonymous input observations $\mathcal{D}$ into $k$ different policies. {\algoname} is inspired by recent advancements in Imitation Learning, world policy, and explainable AI. The Algorithm comprises three major phases: 
% \vspace{0.05in}
\begin{itemize}
\item  We first frame the problem as an IL task with anonymous state-action pairs to learn a unique \textit{world policy} able to emulate all the agent behaviors;
\item We compute local explanations of the \textit{world policy} for each state-action pair through SHAP (SHapley Additive exPlanations). The explanations describe the different agent behaviors upon each observation;
\item We discuss SHAP values properties and how they facilitate the clustering. Finally, we apply a \textit{K-Means} algorithm to group the observations into $k$ policies.

\end{itemize}
% \vspace{-0.05in}
\subsection{World Policy}
Hypothetically we could learn the agent policies by solving $n$ IL problems, one for each agent; however, anonymous state-action pairs restrict this approach due to the lack of individual agent trajectories. 
Another approach could be a world policy $\hat{\pi}_{\theta}$ trained over all the observations $\mathcal{D}$. 
\\
Recent work shows that highly complex environments with multiple autonomous agents can be successfully simulated by learning a unique \textit{world-policy} able to mimic all the agents~\cite{ha2018recurrent,coletta2021towards,coletta2022learning}.
To truthfully simulate the real world and the agent interactions, these models have to learn and impersonate multiple heterogeneous behaviors upon different input from the environment. Therefore, they inherently hold some knowledge about all the agent strategies.

Hereafter, we assume that a \textit{world-policy} $\hat{\pi}_{\theta}$ is the most parsimonious model  able to represent the different agent behaviors by learning their mapping from states to actions from a set of observations $\mathcal{D}$:

\begin{equation}
    \hat{\pi}_{\theta} = \argmin\limits_{\theta} \sum\limits_{(s,a) \in \mathcal{D}} l(\pi_{\theta}(s), a)
\end{equation}
where $l$ is the loss function of the learning problem.
\\
Given the world policy, we can study its behavior over each state-action pair $(s_i, a_i)$, to infer the different strategies adopted by the real agents, and cluster the observations accordingly. In {\algoname} we train a world policy $\hat{\pi}_{\theta}$ in a behavioral cloning fashion, which learns the expert policies using supervised learning.\footnote{Different training procedures or approaches can be applied to learn the \textit{world-policy}, including adversarial training~\cite{ho2016generative}.} 

\subsection{Local explanations}
Explaining the \textit{world-policy} $\hat{\pi}_{\theta}$ behavior under different inputs can reveal details about the strategies adopted by the real agents.
{\algoname} borrows a successful technique from explainable AI, namely SHAP (SHapley Additive exPlanations)~\cite{lundberg2017unified} to disentangle the world-policy's complexity and study its behavior. 

For a demonstration $(s, a)$ we can compute the SHAP values as a vector $\boldsymbol{\phi} \in \mathbb{R}^{|F|}$ defined as the average marginal contribution $\phi_i$ of each state feature $i \in F$ to the prediction $\hat{\pi}_{\theta}(s) \approx a$.
Therefore, we can express the behavior of a real agent for an observation $j$ as the approximated mapping from state $s$ to action $a$, defined by the SHAP values vector $\boldsymbol{\phi}_j$ (which describes how each state feature $i$ drives the agent behavior $a$ in the state $s$). 
In general, the state $s$ can be any representation of the environment in which the agent interacts, including images, text and tabular data, as SHAP values method can deal with complex and continuous data.

% \vspace{-0.1in}
\subsection{{\algoname} Algorithm}
We now describe how {\algoname} combines IL and SHAP values to group the anonymous state-action pairs. The full procedure is shown in Algorithm~\ref{alg:k_shap_algo}.

First, {\algoname} learns the world policy using all the available observations \textbf{(line 1)} in a behavioral cloning fashion. It trains a model using supervised learning, where each state $s$ is used as input and the related action $a$ is considered as the output for the model. Then, by leveraging the world-policy $\hat{\pi}_{\theta}$ and SHAP, it computes the explanations (i.e., SHAP values) for each observation $(s, a) \in \mathcal{D}$. The explanations represent the anonymous observations in a new SHAP values space $\mathcal{D}^{\phi} = \{\boldsymbol{\phi}_i \}_{i=0}^{T}$ \textbf{(lines 3-7)}. Finally, we adopt an existing clustering approach, namely K-Means, to group observations according to their SHAP values, such that observation in the same cluster will be originated from the same strategy \textbf{(line 8)}. Formally, we cluster observations by minimizing the inertia, or within-cluster sum-of-squares, of their SHAP values:
% \vspace{-0.1in}
\begin{equation}
\sum_{j=0}^{k} \sum_{\boldsymbol{\phi}_i \in C_j} ||\boldsymbol{\phi}_i - \mu_j||^2
\end{equation}
where $\boldsymbol{\mu}$ are the cluster centroids, and $C_j$ identify the observations assigned to cluster $j$.
\begin{algorithm}[t]
{
\setstretch{0.8}
\footnotesize
\SetAlgoCaptionSeparator{:}
\KwIn{A set of observations $\mathcal{D} = \{(s_i, a_i)\}_{i=1}^{T}$, the number of clusters $k$, the IL loss function $l$}
\KwOut{Labels $\mathcal{L}$ for each observation to a given policy}
\vspace{0.1in}

% fitting world policy
$\hat{\pi}_{\theta} = \argmin\limits_{\theta} \sum\limits_{(s_i,a_i) \in \mathcal{D}} l(\pi_{\theta}(s_i), a_i)$ \label{algo:world_agent} \\

% SHAP values
$\mathcal{D}^{\phi} = \langle \rangle $\\

\For{$i=0$ \KwTo $|\mathcal{D}|$}
{   
$\boldsymbol{\phi} = \langle \rangle $\\

\ForEach{$j \in F$}
{

\mbox{  $\boldsymbol{\phi}[j] \leftarrow \sum\limits_{S
\subseteq F \setminus {j}} \omega_{S}
    \big[\hat{\pi}_{S \cup \{j\}}(s_{i_{S \cup \{j\}}}) - \hat{\pi}_{S}(s_{i_{S}}) \big]$} \label{algo:explain}\\
    \ \  \ \text{where} $\omega_{S} = \frac{|S|!(|F| - |S| - 1)!}{|F|!} $\\

}

$\mathcal{D}^{\phi}[i] \leftarrow \boldsymbol{\phi} $\\
}

$\mathcal{L}$ = K-Means$(\mathcal{D}^{\phi}, k)$ \label{algo:kmeans}\\

\Return $\mathcal{L}$
\caption{K-SHAP Algorithm}
\label{alg:k_shap_algo}
}
\end{algorithm}%\vspace{-0.1in}

\begin{figure*}[t]
% trim from right edge
%<left> <lower> <right> <upper>}
\includegraphics[trim={0 0.2cm 0 0},width=\linewidth]{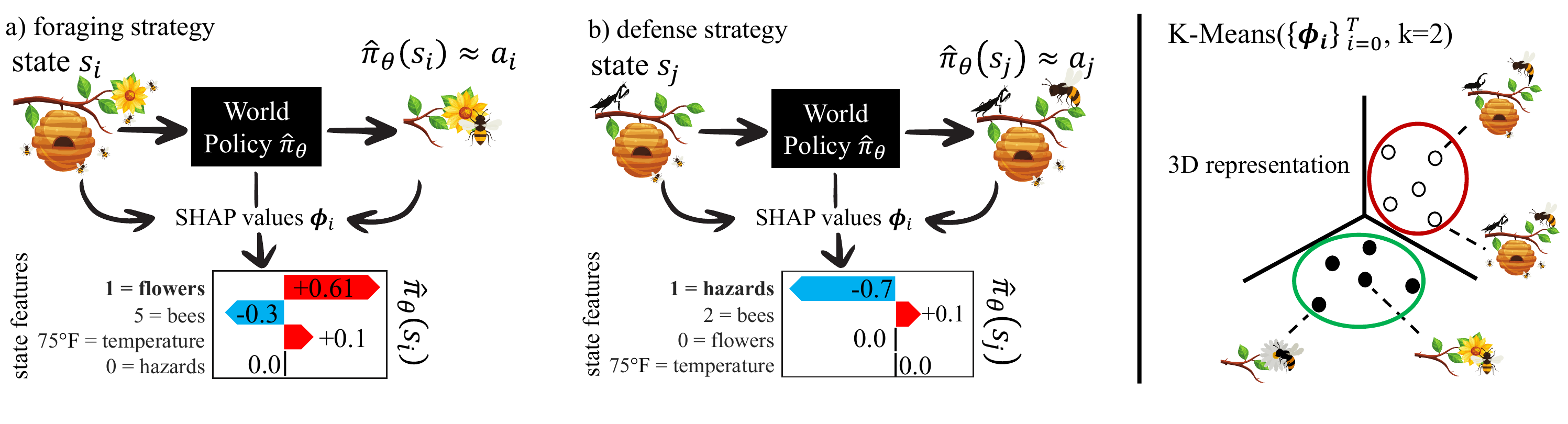}
\centering
\caption{\textbf{Example of {\algoname} clustering. } 
}\label{fig:example-expl}
\vspace{-0.05in}
\end{figure*}

We now briefly discuss two essential properties of SHAP values that we leverage to solve our $k$ policy-clustering problem.
\\
The first implication of the new space $\mathcal{D}^{\phi}$ is the ability to capture the agent behaviors as a response to the environment. In fact, SHAP values explain the world-policy $\hat{\pi}_{\theta}$ predictions with a linear \textit{explanation model} $g$ that captures state features contributions to an action $a$. By assuming that the world-policy mimics the different agent behaviors, the SHAP values define how the agents respond to the different input features. Therefore, the SHAP vectors $\boldsymbol{\phi}$ are similar for observations that originate from comparable behaviors --- they respond similarly to environmental changes (i.e., features) -- and they are close in the vector space $\mathcal{D}^{\phi}$. Contrarily, dissimilar behaviors have different vectors in the space.
\\
Secondly, SHAP values offer a natural advantage in clustering problems~\cite{lundberg2018consistent}: by definition SHAP values convert all the state features into the same metric space. Each feature will have the same units as the model output, as the features measure their impact on the model output in the new SHAP space. This enables us to easily compare and cluster data, regardless of the different magnitude and units of the features. For example, while in Figure \ref{fig:example-expl} it can be challenging to compare the temperature in Fahrenheit to the number of flowers, in the SHAP space both the features will have the same unit, representing their contribution to the action $a$. 

% \vspace{-0.1in}
\paragraph{K-SHAP training}
Our method generally requires selecting the number of clusters k as a unique hyper-parameter. When $k$ is unknown, we adopt the classic Elbow method to optimize $k$ according to the \textit{distortion} or \textit{variance} of the clusters in the SHAP values space. 
In Appendix~\ref{app:k_unkwnn}, we compare the clustering performance in scenarios where the value of $k$ is either unknown or provided. In Section \ref{sec:exp}, we also show that we can use dimensionality reduction techniques like UMAP~\cite{mcinnes2018umap} on the SHAP values space to visualize or improve clustering.  
\\
Finally, as we adopt a supervised IL technique we may need to use multiple \textit{world-policies} in case of observations with identical states to disambiguate them and let the models learn the correct behaviors. In the case of identical states, a unique world-policy may average actions from multiple agents, with inaccurate SHAP values and clustering. 

% \vspace{-0.15in}
\paragraph{Algorithm illustration}
As an illustrative example, in Figure \ref{fig:example-expl} we consider honey bees that act according to two different unknown behaviors (a) foraging and (b) defense. For simplicity, we describe the honey bee action $a \in [-1,1]$ as the probability of staying close to protect the hive ($a \approx -1$) or moving to search for food ($a \approx 1$). We identify four state features describing the number of \textit{flowers}, \textit{bees}, \textit{temperature}, and \textit{hazards} close to the hive. 

$\bullet$ In the scenario (a), we consider a foraging observation in which the honey bee observes the environment and flies close to the flower. We explain the honey behavior through the world policy using SHAP values. The SHAP values describe the honey bee behavior (i.e., search for food) as a strong contribution to the action $a=1$ from the presence of a flower (i.e., food), and a small contribution from presence of warm weather (i.e., temperature). The  presence of other bees around the hive is instead a negative contribution, as they may compete for the flower. % 

$\bullet$  In the scenario (b), we consider a defense observation in which a honey bee observes the environment and chooses to defend the hive. In this case the SHAP values describe the bee behavior as a major contribution to the action $a=-1$ from the presence of one hazard (i.e., mantis). The presence of few bees slightly reduces the need to defend the hive.

Finally, in the leftmost picture we show how we use the computed SHAP values to group the observations into $k=2$ policies through a K-Means algorithm. The picture shows SHAP values from foraging and defense observations that lay into different regions of the space, i.e., clusters.

% \vspace{-0.1in}
\section{Experiments}\label{sec:exp}
% \vspace{-0.1in}
We now experimentally evaluate {\algoname} on both synthetic and real data. We apply K-SHAP to anonymous observations from agents in a financial market (i.e., traders), and we experimentally demonstrate that we can identify state-action pair clusters dictated by the different behavioral strategies.
We average the results over 20 runs, and we assume to know the number of clusters $k$.\footnote{In Appendix \ref{app:k_unkwnn}, we show that {\algoname} achieves similar performance also when the number of clusters $k$ is unknown.
%.
} We consider a Random Forest as the \textit{world-policy} for {\algoname}, and we compute the SHAP values using TreeSHAP~\cite{lundberg2020local}.  We consider two variants of our approach: {\algoname} in which we directly cluster the SHAP values; and {\algoname} ($\mathcal{Z}$) in which we cluster a 2-dimensional embedding $\mathcal{Z}$ of the SHAP values, obtained by applying UMAP~\cite{mcinnes2018umap}. UMAP is a dimensionality reduction technique similar to t-SNE~\cite{van2008visualizing} that preserves more the global structure of the data, and requires less computational resources. 

% \vspace{-0.15in} 
\paragraph{Data}
We first use the state-of-art multi-agent market simulator ABIDES~\cite{byrd2019abides} to simulate synthetic market data. We would like to underline that simulated environment allows to generate state-action pairs with agent IDs, hence, providing the ground truth information for our study. We let market agents realize six distinct trading strategies: \footnote{A detailed description of ABIDES agents is provided in Appendix \ref{app:abides}.} 

$\bullet$ \textit{Market Making}~\cite{chakraborty2011market} that provides liquidity to the market by placing both buy and sell orders; 
\\
$\bullet$  \textit{Fundamental Trading}~\cite{wah2017welfare} that trades according to a belief of the real stock value; 
\\
$\bullet$ \textit{$(\delta_1, \delta_2)$-based Momentum Trading}~\cite{byrd2019abides} that trades following two momentum signals of the price (i.e., moving averages) computed over the last $\delta_1$ and $\delta_2$ minutes; 
\\
$\bullet$ \textit{Noise Trading}~\cite{gode1993allocative} that places orders randomly; 
\\
$\bullet$ \textit{Irrational RL Trading}~\cite{liu2022biased} that models a sub-rational human behavior; 
\\
$\bullet$ \textit{Rational RL Trading}~\cite{liu2022biased} that models an electronic (i.e., rational) trading algorithm.

By combining these strategies, we generate six scenarios:

$\bullet$ \textbf{Abides $\boldsymbol{\pi}^3$} comprises of \textit{Market Making}, \textit{$(12,26)$-based Momentum}, and \textit{Fundamental} trading strategies, with 127 agents. 
\\
$\bullet$ \textbf{Abides $\boldsymbol{\pi}^4$} comprises of \textit{Market Making}, \textit{$(12,26)$-based Momentum}, \textit{Fundamental}, and \textit{Noise} trading strategies, with  5127 agents. 
\\
$\bullet$ \textbf{Abides $\boldsymbol{\pi}^5$} comprises of \textit{Market Making}, \textit{$(0.2,0.4)$-based Momentum}, \textit{$(12,26)$-based Momentum}, \textit{$(48,96)$-based Momentum}, and \textit{Fundamental} trading strategies, with 157 agents. 
\\
$\bullet$ \textbf{Abides $\boldsymbol{\pi}^6$} comprises of \textit{Market Making}, \textit{$(0.2,0.4)$-based Momentum}, \textit{$(12,26)$-based Momentum}, \textit{$(48,96)$-based Momentum}, \textit{Fundamental}, and \textit{Noise} trading strategies, with 5157 agents.
\\
$\bullet$ \textbf{RL-Agents Bubble} comprises of a \textit{Irrational} and \textit{Rational} RL agent, with a \textit{Bubble} market scenario~\cite{siegel2003asset}.
\\
$\bullet$  \textbf{RL-Agents Sine} comprises of a \textit{Irrational} and \textit{Rational} RL agent, with a cyclic market scenario.

In the last two scenarios, we model the market and train the RL agents according to the original paper~\cite{liu2022biased}. In these two scenarios we simulate 11 days of market, while in the first four scenarios we simulate 4 days.

For real market data we consider historical data from NASDAQ stock exchange where agent IDs are not available~\cite{totalview}. We consider three stocks (i.e., AVXL, AINV, and ADAP) over 4 trading days from 05$^{th}$ to 8$^{th}$ Jan 2021.

\paragraph{Metrics}
For synthetic data the ground-truth cluster labels are available, and we apply three standard metrics to evaluate the clustering: \textit{purity score}, \textit{Adjusted Rand Index} (ARI)~\cite{hubert1985comparing} and \textit{Normalized Mutual Information} (NMI)~\cite{vinh2009information}. The purity score ranges from 0 to 1, and it evaluates how homogeneous each cluster is (where 1 being a cluster consists of observations from a single strategy). The ARI ranges from -1 to 1, and it represents the \textit{adjusted for chance} version of Rand index, which measures the percentage of correct cluster assignments (where 1 being a perfect clustering and 0 being a random clustering). The NMI ranges from 0 to 1 (where 1 indicated a perfect clustering), and it measures how much information is shared between the clusters and the labels, adjusted by the number of clusters. 
\\
For the historical market data where the ground truth is not available we evaluate the \textit{Silhouette Index}~\cite{rousseeuw1987silhouettes} and we introduce a \textit{Utility} metric for the clusters. The \textit{Silhouette Index} measures the similarity of each state-action pair to its own cluster compared to other clusters. It ranges between -1 and 1, where 1 indicate the highest degree of confidence that the observation belongs to a correct cluster. The \textit{Utility} evaluates the learning improvement when we learn a unique policy from all the state-action pairs w.r.t. learning $k$ policies from the identified clusters. Thus, an improvement in \textit{Utility} indicates that each cluster contains homogeneous state-action pairs in terms of strategy, as a model trained on them can predict the next action more easily. An higher \textit{Utility} indicates a better clustering. Both the Silhouette Index and the \textit{Utility} are the appropriate metrics to use as they can validate the cluster analysis in the absence of ground truth.\footnote{We provide a mathematical definition of the metrics in Appendix \ref{app:metrics}.} 

\vspace{-0.15in}
\paragraph{Benchmarks} To the best of our knowledge this is the first approach to tackle policy clustering under anonymous state-action pairs. We first compare our approach against data clustering methods, which represent a natural first attempt to cluster anonymous observations: 
\\
$\bullet$ We consider \textit{K-Means}~\cite{hartigan1979algorithm} to group the anonymous observations, using the state-action pairs directly (K-Means), or their 2d embedding $\mathcal{Z}$ obtained through UMAP (K-Means ($\mathcal{Z}$)).
\\
$\bullet$ We consider Deep Clustering Network (DCN)~\cite{mukherjee2019clustergan} that jointly optimizes dimensionality reduction and clustering. The dimensionality reduction is accomplished via learning a deep autoencoder. We apply DCN on the anonymous state-action pairs.
\\
$\bullet$ We consider ClusterGAN~\cite{mukherjee2019clustergan} that clusters by back-projecting the data to the latent-space learned by a generative adversarial network (GAN)~\cite{goodfellow2020generative}. It introduces a mixture of one-hot and continuous variables as latent variables, which retain information about the data and reveal some intrinsic clustering structure. 
{\color{black}
\\
$\bullet$ We also consider HC-MGAN~\cite{de2022top}, a Hierarchical Clustering approach using Multiple GANs to implicitly learn a latent representation of the data. In particular, this work clusters data by exploiting the fact that each GAN tends to generate data that correlates with a sub-region of the real data distribution.
}
\\
Finally, we consider $\Sigma$-GIRL~\cite{ramponi2020truly}, a recent work that solves an intention-clustering problem. This work clusters agents based on their goals in an expectation-maximization (EM) fashion, using IRL in the maximization step. It solves a problem similar to ours, but it does not consider anonymous state-action pairs. We consider a modified version called \textit{EM K-Clustering}. Instead of a joint optimization process in which we learn the agent-cluster assignments and the reward functions through IRL, we learn the observation-cluster assignments and the policy functions through IL. We consider a Neural Network (NN) to learn the policies through IL.

\begin{figure}[t]
% <left> <lower> <right> <upper>}
\includegraphics[trim={0 0 0 0},width=\linewidth]{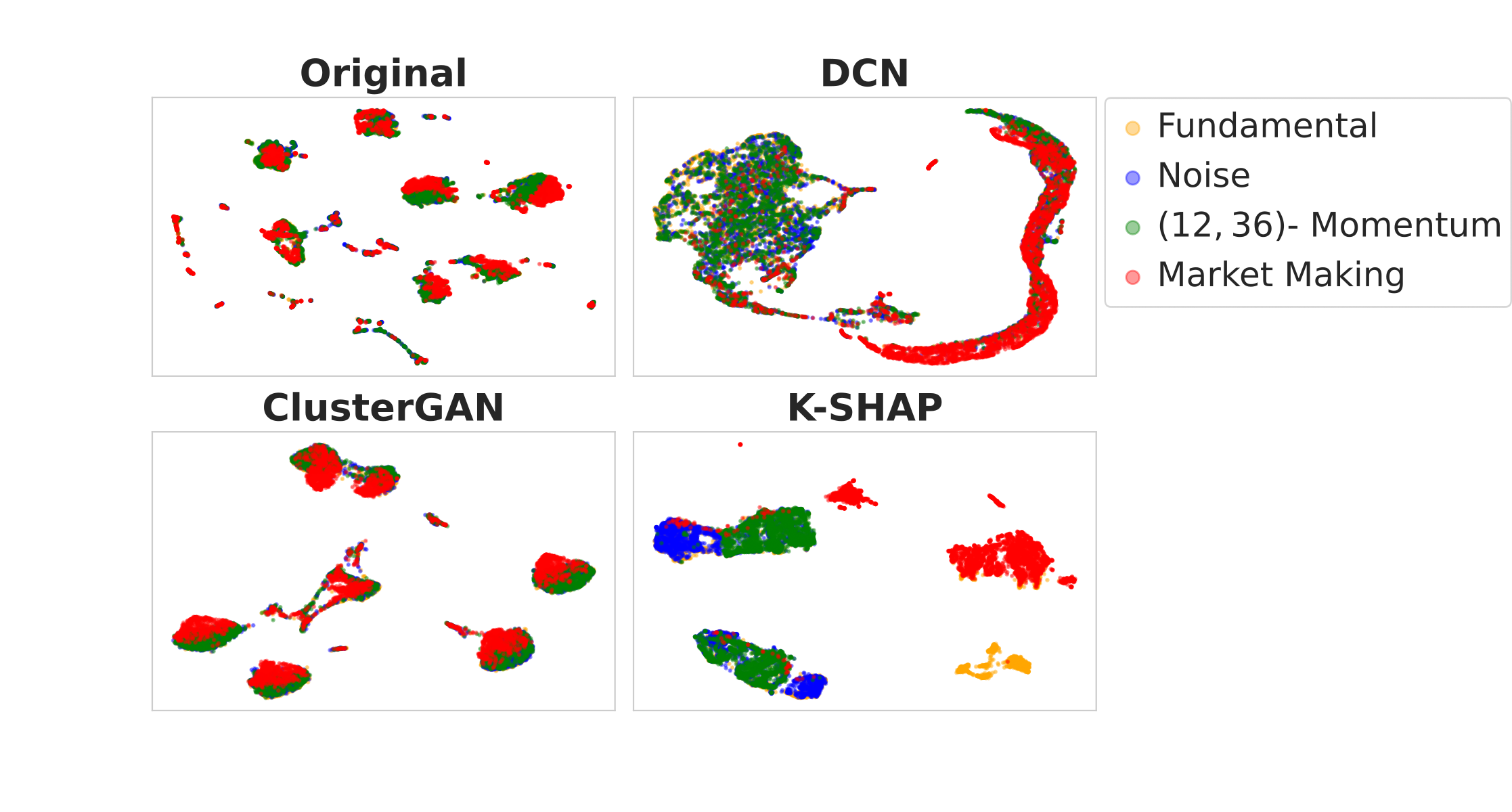}
\caption{Two-dimensional space visualization of state-action pairs and learned clustering embedding, using UMAP. K-SHAP better reveals the intrinsic clustering structure among data.}\label{fig:shap_space}
    % \vspace{-0.1in}
         \vspace{-0.1in}
\centering
\end{figure}

\subsection{Multi-Agent Synthetic Data} 

Table \ref{tab:abides_perf} investigates the algorithm performance using the simulated market data with ground truth labels. In all the scenarios, our algorithm {\algoname} outperforms existing work. In \textbf{Abides $\boldsymbol{\pi}^3$}, {\algoname} outperforms other approaches of around 300\% in terms of ARI and NMI, while it consistently and significantly shows the best performance in all \textbf{Abides} scenarios. In the scenarios with two RL agents, most of the approaches fail to distinguish between the two different strategies. However, when we apply {\algoname} using a 2-dimensional embedding of the SHAP values (K-SHAP ($\mathcal{Z}$)) we obtain encouraging results. It achieves 10 times better performance in terms of ARI and NMI. It is worth noticing that the purity score is less relevant with only two clusters (e.g., if we put all the observations in one cluster, we still obtain a purity score of $0.5$).
\\
Figure \ref{fig:shap_space} shows a 2-dimensional visualization of the original state-action pairs against the learned clustering embedding for DCN, ClusterGAN, and {\algoname}. We omit K-Means and EM K-Clustering, as they do not use embedding representations. We consider \textit{Abides $\boldsymbol{\pi}^4$} and the colors represent the ground-truth labels. The figure clearly shows how the original state-action pairs do not retain any natural structure to help the clustering. Moreover, ClusterGAN latent space does not efficiently cluster the observations, while DCN only partially identifies Market Makers observations. Instead, for our algorithm the SHAP values better reveal the intrinsic clustering structure among data, and only Noise Agents, which place random orders, are slightly confused with Momentum Agents.

\begin{table}[]
    \centering
\resizebox{\columnwidth}{!}{%
\begin{tabular}{lllll}
\toprule
Dataset & Algorithm &        ARI &        NMI &     Purity \\
\midrule
\multirow{7}{*}{\begin{tabular}[c]{@{}l@{}}
\textbf{Abides $\boldsymbol{\pi}^3$}\\ \cite{byrd2019abides} \end{tabular}}  & K-Means  &  0.00±0.00 &  0.00±0.00 &  0.35±0.00 \\
 & K-Means ($\mathcal{Z}$) &  0.00±0.00 &  0.00±0.00 &  0.35±0.00 \\
         & ClusterGAN~\cite{mukherjee2019clustergan} &  0.11±0.10 &  0.14±0.14 &  0.33±0.25 \\
                & DCN~\cite{yang2017towards} &  0.10±0.09 &  0.14±0.12 &  0.50±0.08 \\
                & HC-MGAN~\cite{de2022top} &  0.03±0.01 &  0.07±0.02 &  0.42±0.02 \\
                & EM K-Clustering &  0.13±0.07 &  0.15±0.11 &  0.53±0.07 \\
                & K-SHAP &  \textbf{0.50±0.10} &  \textbf{0.56±0.05} &  \textbf{0.77±0.03} \\
                & K-SHAP ($\mathcal{Z}$) &  0.23±0.00 &  0.25±0.00 &  0.60±0.00 \\ \hline
\multirow{7}{*}{\begin{tabular}[c]{@{}l@{}}
\textbf{Abides $\boldsymbol{\pi}^4$}\\ \cite{byrd2019abides}  \end{tabular}} & K-Means  &  0.00±0.00 &  0.00±0.00 &  0.27±0.01 \\  
& K-Means ($\mathcal{Z}$) &  0.00±0.00 &  0.00±0.00 &  0.28±0.00 \\
                & ClusterGAN~\cite{mukherjee2019clustergan} &  0.08±0.08 &  0.11±0.12 &  0.24±0.19 \\
                & DCN~\cite{yang2017towards} &  0.04±0.06 &  0.06±0.04 &  0.33±0.07 \\
                & HC-MGAN~\cite{de2022top} &  0.04±0.01 &  0.07±0.01 &  0.36±0.02 \\
                & EM K-Clustering &  0.08±0.04 &  0.11±0.06 &  0.43±0.10 \\
                & K-SHAP &  \textbf{0.35±0.07} &  \textbf{0.52±0.02} &  \textbf{0.65±0.11} \\
                & K-SHAP ($\mathcal{Z}$) &  0.21±0.08 &  0.32±0.03 &  0.55±0.03 \\ \hline
\multirow{7}{*}{\begin{tabular}[c]{@{}l@{}}
\textbf{Abides $\boldsymbol{\pi}^5$}\\ \cite{byrd2019abides} \end{tabular}}  & K-Means  &  0.00±0.00 &  0.01±0.00 &  0.23±0.00 \\
               & K-Means ($\mathcal{Z}$) &  0.01±0.00 &  0.01±0.00 &  0.24±0.00 \\
                & ClusterGAN~\cite{mukherjee2019clustergan} &  0.07±0.06 &  0.10±0.08 &  0.22±0.17 \\
                & DCN~\cite{yang2017towards} &  0.08±0.04 &  0.11±0.05 &  0.33±0.04 \\
                & HC-MGAN~\cite{de2022top} &  0.04±0.01 &  0.07±0.00 &  0.29±0.01 \\ 
                & EM K-Clustering &  0.07±0.03 &  0.12±0.06 &  0.40±0.12 \\
                & K-SHAP &  \textbf{0.22±0.08} &  \textbf{0.37±0.06} &  \textbf{0.51±0.10} \\
                & K-SHAP ($\mathcal{Z}$) &  0.20±0.03 &  0.33±0.01 &  0.50±0.06 \\ \hline
\multirow{7}{*}{\begin{tabular}[c]{@{}l@{}}
\textbf{Abides $\boldsymbol{\pi}^6$}\\ \cite{byrd2019abides} \end{tabular}}  & K-Means  &  0.00±0.00 &  0.01±0.00 &  0.20±0.01 \\
                & K-Means ($\mathcal{Z}$) &  0.01±0.00 &  0.01±0.00 &  0.21±0.00 \\
                & ClusterGAN~\cite{mukherjee2019clustergan} &  0.06±0.04 &  0.10±0.07 &  0.19±0.15 \\
                & DCN~\cite{yang2017towards} &  0.06±0.03 &  0.11±0.07 &  0.29±0.04 \\
                & HC-MGAN~\cite{de2022top} &  0.02±0.00 &  0.05±0.01 &  0.23±0.01 \\
                & EM K-Clustering &  0.07±0.03 &  0.12±0.04 &  0.38±0.15 \\
                & K-SHAP &  \textbf{0.15±0.01} & \textbf{0.31±0.01} &  \textbf{0.42±0.01} \\
                & K-SHAP ($\mathcal{Z}$) &  0.15±0.02 &  0.29±0.01 &  0.42±0.02 \\ \hline
\multirow{7}{*}{\begin{tabular}[c]{@{}l@{}}
\textbf{RL-Agents Bubble} \\ \cite{liu2022biased} \end{tabular}}  & K-Means  &  0.00±0.00 &  0.00±0.00 &  0.58±0.00 \\
             & K-Means ($\mathcal{Z}$) &   0.00±0.00 &  0.00±0.00 &  0.58±0.00 \\
                & ClusterGAN~\cite{mukherjee2019clustergan} &   0.00±0.00 &  0.00±0.00 &  0.37±0.27 \\
                & DCN~\cite{yang2017towards} &  0.00±0.00 &  0.00±0.00 &  0.58±0.00 \\
                & HC-MGAN~\cite{de2022top} &  0.00±0.00 &  0.00±0.00 &  0.58±0.00 \\ 
                & EM K-Clustering &   0.00±0.00 &  0.00±0.00 &  0.58±0.00 \\
                & K-SHAP &  0.00±0.00 &  0.00±0.00 &  0.58±0.00 \\
                & K-SHAP ($\mathcal{Z}$) &   \textbf{0.42±0.20} &  \textbf{0.35±0.15} &  \textbf{0.81±0.10} \\ \hline
\multirow{7}{*}{\begin{tabular}[c]{@{}l@{}}
\textbf{RL-Agents Sine} \\ \cite{liu2022biased} \end{tabular}} & K-Means  &   0.00±0.00 &  0.00±0.00 &  0.56±0.00 \\
                & K-Means ($\mathcal{Z}$) &  0.00±0.00 &  0.00±0.00 &  0.56±0.00 \\
                & ClusterGAN~\cite{mukherjee2019clustergan} &   0.00±0.00 &  0.00±0.00 &  0.39±0.28 \\
                & DCN~\cite{yang2017towards} &   0.00±0.00 &  0.00±0.00 &  0.56±0.00 \\
                & HC-MGAN~\cite{de2022top} &   0.00±0.00 &  0.01±0.00 &  0.58±0.01 \\
                & EM K-Clustering &    0.02±0.00 &   0.01±0.00 &   0.58±0.00 \\
                & K-SHAP &   0.02±0.00 &  0.01±0.00 &  0.56±0.00 \\
                & K-SHAP ($\mathcal{Z}$) &   \textbf{0.25±0.07} &  \textbf{0.20±0.06} &  \textbf{0.75±0.04} \\ 
\bottomrule
\end{tabular}}    
\caption{Multi-Agent Synthetic Data}
    \label{tab:abides_perf}
 \vspace{-0.3in}
\end{table}

\begin{table}[t]
    \centering
\resizebox{\columnwidth}{!}{%
\tiny
\begin{tabular}{llll}
\toprule
Stock & Algorithm &  Utility & Silhouette    \\
\midrule
\textbf{ADAP} & K-Means  &  0.00±0.00 &   \textbf{0.23±0.01} \\
  \cite{totalview}   & K-Means ($\mathcal{Z}$) &   0.06±0.01 &  0.00±0.01 \\
     & ClusterGAN~\cite{mukherjee2019clustergan} &   0.12±0.11 &   0.02±0.01 \\
     & DCN~\cite{yang2017towards} &  0.00±0.00 &   0.00±0.00 \\
     & EM K-Clustering &   \textbf{0.47±0.02} &   0.01±0.01 \\
     & K-SHAP &   0.33±0.00 &   0.12±0.00 \\
     & K-SHAP ($\mathcal{Z}$) &   0.21±0.04 &   0.00±0.01 \\ \hline
\textbf{AINV} & K-Means  &   0.00±0.00 &   \textbf{0.24±0.00} \\
  \cite{totalview}   & K-Means ($\mathcal{Z}$) &   0.05±0.02 &  0.00±0.00 \\
     & ClusterGAN~\cite{mukherjee2019clustergan} &   0.07±0.06 &   0.01±0.01 \\
     & DCN~\cite{yang2017towards} &  0.00±0.00 &  0.00±0.01 \\
     & EM K-Clustering &   \textbf{0.43±0.02} &   0.01±0.03 \\
     & K-SHAP &   0.19±0.06 &   0.11±0.06 \\
     & K-SHAP ($\mathcal{Z}$) &   0.10±0.05 &   0.01±0.01 \\ \hline
\textbf{AVXL} & K-Means  &  0.00±0.01 &  \textbf{0.21±0.00} \\
  \cite{totalview}   & K-Means ($\mathcal{Z}$) &   0.05±0.04 &   0.12±0.04 \\
     & ClusterGAN~\cite{mukherjee2019clustergan} &   0.13±0.03 &   0.01±0.01 \\
     & DCN~\cite{yang2017towards} & 0.14±0.04 &  0.02±0.02 \\
     & EM K-Clustering &   \textbf{0.53±0.04} &   0.01±0.02 \\
     & K-SHAP &   0.36±0.01 &   0.09±0.00 \\
     & K-SHAP ($\mathcal{Z}$) &   0.24±0.05 &   0.02±0.03 \\
\bottomrule
\end{tabular}}
\caption{Historical Market Data}
    \label{tab:real_stock_dataset}
     \vspace{-0.2in}
\end{table}

 \vspace{-0.1in}
\subsection{Historical Market Data}
% \vspace{-0.1in}
Finally, we consider an experiment with historical market data from NASDAQ stock exchange where agent IDs are not available.  
We define the Utility as the percentage difference between the mean squared error (MSE) obtained by a NN trained over all the state-action pairs, and the MSE obtained by $k$ NNs trained over the identified clusters. Each NN predicts the next action $a$ given the current market state $s$, and all the NNs have the same architecture.

Table \ref{tab:real_stock_dataset} shows that \textit{K-Means} always achieves the best Silhouette while the best algorithm for the Utility is \textit{EM K-Clustering}. However, \textit{K-Means} optimizes the clusters for within-cluster sum-of-squares which maximizes the Silhouette; while \textit{EM K-Clustering} optimizes the clusters using $k$ policies in an EM fashion, which indirectly maximizes the Utility metric. Both the approaches easily succeed in these metrics but they fail short in the other metric: \textit{K-Means} has 0 Utility, and \textit{EM K-Clustering} has 0.01 Silhouette. Instead, {\algoname} achieves a good trade-off between Silhouette and Utility, without maximizing any of them directly.  
In all the scenarios we consider $k=3$, we refer to Appendix \ref{app:vary_k} for more experiments and charts.

{\color{black}
\subsection{The impact of state features}
While K-SHAP is most effective when the latent policies use multiple state features to make decisions, K-SHAP can also distinguish strategies using only few features. Given the local accuracy property (see Section~\ref{sec:shap}), the local explanation $\phi_i$ for a state feature $i$ can be both negative or positive, with different magnitudes according to the impact on the action. Therefore, policies that use the same few features, but in different ways, can still be distinguished; while with null states our approach may fail. To better study the applicability of our approach with few and no features, we created an iterated prisoner’s dilemma and a one-shot prison dilemma.

First, we consider a strategy that always ``betrays'' and one that always ``cooperates''. K-SHAP achieves $Purity = 0.5$, NMI = $0$ and ARI = $0$ (i.e., random clusters) when the state is null. It achieves $Purity$=NMI=ARI=$1$ (i.e., perfect clusters) when the state is provided (i.e., time-step and previous opponent’s action). In the latter scenario, with just a two-dimensional state space, the SHAP latent space is meaningful and can be used to distinguish the policies.

As a further example we consider a strategy that always ``cooperates'' and a strategy that flips his behavior w.r.t. the previous action.  When the state is null we have random clusters. When the state is provided (i.e., time-step and previous action)  K-SHAP achieves $Purity=0.75$, $NMI=0.35$, and $ARI=0.25$. We introduced this example to highlight that: while the SHAP values $\boldsymbol{\phi}$ can be useful even when the strategies rely on the same few features, the world policy may underfit with such a small state space and overlapping agent strategies, resulting in worse clusters. 

We conclude that the number of state features, and how the latent policies use them, is important for both the SHAP values and the world policy. 
}

% \vspace{-0.1in}
\section{Conclusions and Future Work}
% \vspace{-0.1in}
We proposed a Policy Clustering algorithm, {\algoname}, that groups anonymous state-action pairs according to the agent strategies. We framed the problem as an IL to learn a compact representation of the agent strategies as a world-policy; we explained such a policy using SHAP; finally, we use the SHAP values to group observations according to the agent behaviors. We shown that {\algoname} consistently outperforms existing work on both synthetic and real market data.
As we only access anonymous observations, the current approach works well when the actions are mostly driven by the state features: {\algoname} has not enough information to classify an action driven only by a sequence of previous actions, unless the state retains traces of the agent actions. Similarly, identical state-action pairs cannot be classified with different strategies. As future work, we envisioned to further explore and apply {\algoname} in these scenarios. {\color{black}Please refer to the appendix for more detailed experiments and discussions on our work impact.}

\vspace{-0.1in}
\section*{Disclaimer} 
This paper was prepared for informational purposes by the Artificial Intelligence Research group of JPMorgan Chase \& Co. and its affiliates (``J.P. Morgan''), and is not a product of the Research Department of J.P. Morgan. J.P. Morgan makes no representation and warranty whatsoever and disclaims all liability, for the completeness, accuracy or reliability of the information contained herein. This document is not intended as investment research or investment advice, or a recommendation, offer or solicitation for the purchase or sale of any security, financial instrument, financial product or service, or to be used in any way for evaluating the merits of participating in any transaction, and shall not constitute a solicitation under any jurisdiction or to any person, if such solicitation under such jurisdiction or to such person would be unlawful.

 \vspace{-0.1in}
\section*{Acknowledgements} 
Images are in part designed by Macrovector / Freepik.

\bibliographystyle{named}
\bibliography{main}

\newpage
\appendix
\onecolumn

{\color{black}
\section{Broader Impact}
Our paper introduces a first approach to understand agent strategies and their interaction in anonymized systems. In general, we believe this is an important research area to develop more accurate computational models of complex systems~\cite{goldstone2005computational}, resulting in a better understanding of agent behaviors needed to design more suitable solutions~\cite{bak2021stewardship}.  

For example, in financial markets, identifying archetypal behavioral strategies allows to construct market simulation environments that can reproduce important structural properties of the market~\cite{lebaron2006agent}. Experimentation with counterfactual scenarios in such simulators has been documented to provide significant benefits for the financial community. For instance:
\begin{itemize}
    \item Agent-based simulation that was constructed with similar archetypal agents was used to replay May 2010 flash crash scenario -- hence, promoting the understanding of the flash crash and allowing to subsequently design the policies that prevent flash crashes in the future~\cite{kirilenko2017flash}.
    \item Another benefit of our approach to financial systems is the investigation of market mechanisms that can make markets more fair -- it was demonstrated in simulated market environments that stock exchanges can implement the dynamic fee policy in order to enable equitability of outcomes to market participants~\cite{dwarakanath2022equitable}.
    \item Additionally, evaluating trading strategies or assumptions against poorly calibrated agent-based models can lead to harmful and misleading conclusions (e.g., a severe market crash in 1987 causing \$1.71 trillion losses is attributed to the prevalence of simplistic market models)~\cite{bouchaud2018trades} - hence, it is imperative to build reliable market simulation models for the overall market stability.
\end{itemize}
}

\section{Details of the Datasets}
Here we introduce experimental details 
that do not fit into the main body of the paper. 
\vspace{0.05in}
\\
In our experiments, we consider a stock market in which multiple agents (i.e., traders) interact by selling and buying shares. We consider a synthetic simulated market, which serves as ground truth, and historical market data from NASDAQ stock exchange~\cite{totalview}.
Each state-action pair is generated by a trader in the market. The state represents the recent and ongoing stock market state (e.g., the price, volume and volatility of the stock) and the action is the agent trade (e.g., buy/sell a given number of shares at price $x$). In the following, we discuss in detail how we construct the synthetic datasets, and we summarize the main statistics of both synthetic and historical market data. 

\subsection{Multi-Agent Synthetic data}\label{app:abides} All the synthetic datasets have been generated using a state-of-art multi-agent simulator called ABIDES~\cite{byrd2019abides}, and its 
OpenAI Gym extension called ABIDES-gym~\cite{amrouni2021abides}. The simulator is written in Python3, and it is publicly available at \url{https://github.com/jpmorganchase/abides-jpmc-public}. 
It is a high-fidelity multi-agent market simulation used by practitioners and researchers to generate synthetic markets. The market is generated by simulating the interactions between a given set of agents, for which we can define the number, type, and strategy to simulate different markets. More details on the simulator can be found in the original ABIDES paper~\cite{byrd2019abides}.

\vspace{-0.1in}
\paragraph{Agent Strategies} Here we describe the agent strategies in detail:
\vspace{-0.1in}
\begin{itemize}
    \item \textit{Market Making} agents provide liquidity to the market by both buying and selling shares while making a profit by keeping a low net position. We use the implementation provided in the public repository of ABIDES, (i.e., \texttt{adaptive\_market\_maker\_agent.py}) that follows the Chakraborty-Kearns `ladder' market-making strategy~\cite{chakraborty2011market}, wherein the size of orders placed at each level is set as a fraction of measured transacted volume in the previous time period.
    
    \item \textit{Noise} agents belong to the class of Zero Intelligence (ZI) agents introduced in~\cite{gode1993allocative} to model agents that do not base their trading decisions on the knowledge of market microstructure. \textit{Noise} agents wake up once in the trading day, and place one order randomly. Both their order direction $d \in \{buy, sell\}$, and volume $v \in [1, 100]$ are randomly sampled, while the order price is the best available on the market (i.e., near touch). We use the implementation provided in the public repository of ABIDES, i.e., \texttt{noise\_agent.py}.  
    
    \item \textit{Fundamental Trading} agents belong to the class of ZI agents, but have access to an exogenous fundamental value of the stock, which represents the agent's understanding of the outside world (e.g., assets and earnings of the company). These agents trade the stock according this estimated exogenous fundamental value. They believe that if the stock is overpriced w.r.t. to the fundamental value, the price will go down, and vice versa. Therefore, they sell in the first case and buy in the other one. All the agents have a noisy observation of a simulated fundamental value of the stock, which they use to generate orders. They randomly sample the order volume $v \in [1, 100]$, while the price is the best available for $90\%$ of the orders (i.e., near touch). For $10\%$ of the orders, they use a more aggressive price close to the far touch. We use the implementation provided in the public repository of ABIDES, i.e., \texttt{value\_agent.py}.
    
    \item \textit{$(\delta_1, \delta_2)$-based Momentum Trading} agents trade according to a price momentum indicator of the stock. These agents consider two moving averages (MAs) of the stock price. The \textit{short} MA is computed using the price values in the previous $\delta_1$ minutes, while the \textit{long} MA considers $\delta_2$ minutes. This agent attempts to exploit extreme short-term price moves, by playing a \textit{buy} order when the \textit{short} MA $\delta_1$ $\geq$  \textit{long} MA $\delta_2$, and a \textit{sell} order otherwise. The order volume $v \in [1, 100]$ is randomly sampled, and the price is the best available (i.e., near touch). We extended the implementation provided in the public repository of ABIDES, (i.e., \texttt{momentum\_agent.py}) by changing the MA windows. 

    \item \textit{Irrational RL Trading} agents model a sub-rational human behavior. We train these RL agents according to the original paper~\cite{liu2022biased}, in which the authors model a myopic human investor by decreasing the discount factor $\gamma$ in the Bellman equation~\cite{sutton2018reinforcement}. Therefore, as $\gamma \rightarrow 0$ the RL agent becomes more myopic and trades to maximize only a one-step reward. In the experiments the Irrational RL agents use $\gamma = 0.01$.    
    \item \textit{Rational RL Trading} agents~\cite{liu2022biased} model an electronic (i.e., rational) trading algorithm. We train these RL agents according to the original paper~\cite{liu2022biased}, in which the authors increase the discount $\gamma$ to have a fully rational agent that considers both short-term and long-term rewards. In the experiments the Rational RL agents use $\gamma = 0.99$.
\end{itemize}

\paragraph{Market Scenarios} We combine the agent strategies to generate six different market scenarios.

$\bullet$ In the first four scenarios (Abides $\pi^3$, Abides $\pi^4$, Abides $\pi^5$, and Abides $\pi^6$) we simulate 4 days of synthetic market, considering for each strategy the following number of agents: $5000$ Noise agents; $110$ Fundamental Agents; $2$ Market Making Agents; $15$ (0.2,0.4)-based Momentum Agents; $15$ (12,26)-based Momentum Agents; and $15$ (48,96)-based Momentum Agents.
The agents trade a synthetic stock priced at around $100\$$. The market is simulated with nanosecond time resolution, and we consider 29 state features and 3 features for the agent actions (i.e., order depth, volume, and direction). The detailed statistics for these datasets are reported in Table \ref{tab:abides3} for Abides $\pi^3$; Table \ref{tab:abides4} for Abides $\pi^4$; Table \ref{tab:abides5} for Abides $\pi^5$ and Table \ref{tab:abides6} for Abides $\pi^6$. We use the implementation provided in the public repository of ABIDES (i.e., \texttt{rmsc04.py}) to define the agent interactions.

$\bullet$ The last two scenarios consider RL-based agents, which are simulated according to the two original paper~\cite{liu2022biased}. In particular we consider a \textit{Bubble}~\cite{siegel2003asset} market scenario, and a market scenario in which the price follows a sine wave, with the same starting and closing price. In these scenarios, the market is simulated every minute, and the agents have 9 different actions modeled as an integer $a \in [-4, 4]$. The action $a \in \{-4, -3, -2, -1\}$ represents a buy order of size 2, where the value of $a$ represents the order price w.r.t. the mid-price. The action $a = 0$ represents the \textit{HOLD} action. The action $a \in \{1, 2, 3, 4\}$ represents a sell order of size 2, where the value of $a$ represents the order price w.r.t. the mid-price.  In these scenarios, we simulate 11 days of market data, we consider 20 state features and 1 integer feature for the agent actions. In Table \ref{tab:irr_bubble} and Table \ref{tab:irr_sine} we summarize the dataset properties for the  \textit{Bubble} and \textit{Sine} market scenarios, respectively.

\subsection{Historical Market Data}
Finally, we consider real market data from NASDAQ stock exchange~\cite{totalview}. In particular, we use historical market data that includes all the orders submitted to the market, without any agent identifiers. We consider three stocks (i.e., AVXL, AINV, and ADAP) over 4 days from 05$^{th}$ to 8$^{th}$ Jan 2021. The historical data contains anonymous state-action pairs at  nanosecond time resolution, and we consider the same 29 state features and 3 action features as in \textbf{Abides} synthetic dataset. Table  \ref{tab:ainv}, Table \ref{tab:avxl}, and Table \ref{tab:adap}, summarize the dataset properties for  AINV, AVXL, and ADAP, respectively. 

\section{Algorithm and Benchmark details}

$\bullet$ We consider \textit{K-Means}~\cite{hartigan1979algorithm} to group the anonymous observations, using  the state-action pairs directly (K-Means), or their 2d embedding $\mathcal{Z}$ obtained through UMAP (K-Means ($\mathcal{Z}$)).
We use K-Means implementation provided by scikit-learn~\cite{scikit-learn}, and the official UMAP implementation~\cite{mcinnes2018umap-software}.

$\bullet$ We consider Deep Clustering Network (DCN)~\cite{mukherjee2019clustergan} that jointly optimizes dimensionality reduction and clustering. The dimensionality reduction is accomplished via learning a deep autoencoder. We apply DCN on the anonymous state-action pairs: we feed both the states and the actions to the deep autoencoder that learns their encoding and reconstruction.
We consider the official implementation of the paper available at \url{https://github.com/boyangumn/DCN-New}.

$\bullet$ We consider ClusterGAN~\cite{mukherjee2019clustergan} that clusters by back-projecting the data to the latent-space. It introduces a mixture of one-hot and continuous variables as latent variables, which retain information about the data and reveal some intrinsic clustering structure. We apply ClusterGAN to the anonymous state-action pairs.  
We consider the official implementation of the paper available at \url{https://github.com/sudiptodip15/ClusterGAN}.

$\bullet$ Finally, we consider a modified version of $\Sigma$-GIRL~\cite{ramponi2020truly}. This work solves the intent-clustering problem in an expectation-maximization (EM) fashion, using IRL in the maximization step. Instead of a joint optimization process in which we learn the agent-cluster assignments and the reward functions through IRL, we learn the observation-cluster assignments and the policy functions through IL. In particular, we learn to assign the anonymous observations to $k$ clusters while learning $k$ policies through IL. We consider a Neural Network (NN) to learn the policies through IL. The NN architecture is described in Appendix~\ref{app:nn_architecture}. We modified the official implementation of the paper available at \url{https://github.com/gioramponi/sigma-girl-MIIRL}.

\paragraph{Neural Network Architecture}\label{app:nn_architecture}
Here we briefly describe the architecture of the NN used in the experiments: to solve the IL task in \textit{EM K-Clustering}; to model the world-policy in Appendix \ref{app:wpolicies}; and to compute the Utility score \ref{app:metrics}.
We consider a feedforward Neural Network with 2 linear hidden layers with \textit{Leaky ReLU} activation function, and respectively 64 and 32 neurons. After Each hidden layer we consider a $0.1$-dropout layer. 

\paragraph{World-Policy Architecture}
For the Random Forest world-policy we use the implementation provided by scikit-learn~\cite{scikit-learn}, where we fix the number of trees to 100 and we use mean squared error as objective. For UMAP we use the official implementation~\cite{mcinnes2018umap-software}, and we fix the number of neighbors observations to 15.

\subsection{Metrics}\label{app:metrics}

$\bullet$ The \textit{purity score} ranges from 0 to 1, and it evaluates how homogeneous each cluster is (where 1 being a cluster consists of observations from a single strategy). The purity is defined as follows:
$$
\text{purity}(\Omega, C) = \frac{1}{N} \sum_{k} \max_{j} |\Omega_k \cap C_j |   
$$
where $N$ is the number of observations, $k$ and $j$ are the number of clusters and strategies (ground truth), respectively. We denote with $\Omega$ the set of identified clusters, while $C$ represents the ground truth clusters.

$\bullet$ The \textit{Adjusted Rand Index} (ARI)~\cite{hubert1985comparing} ranges from -1 to 1, and it represents the \textit{adjusted for chance} version of Rand index, which measures the percentage of correct cluster assignments (where 1 being a perfect clustering and 0 being a random clustering). We use the ARI implementation provided by scikit-learn~\cite{scikit-learn}, and we refer to the original paper for further details~\cite{hubert1985comparing}.

$\bullet$ The \textit{Normalized Mutual Information} (NMI)~\cite{vinh2009information} ranges from 0 to 1 (where 1 indicated a perfect clustering), and it measures how much information is shared between the clusters and the labels, adjusted by the number of clusters. We use NMI implementation provided by scikit-learn~\cite{scikit-learn}, and we refer to the original paper for further details~\cite{vinh2009information}.

$\bullet $ The \textit{Silhouette Index} (SI)~\cite{rousseeuw1987silhouettes} measures the similarity of each state-action pair to its own cluster compared to other clusters. It ranges between -1 and 1, where 1 indicate the highest degree of confidence that the observation belongs to a correct cluster. Let $b_i$ be the mean euclidean distance between a sample $i$ and all other points in the same cluster; and $c_i$ the mean distance between the sample $i$ and all the other points in the nearest cluster, then the SI is computed as follows:
$$
SI = \frac{1}{N} \sum_{i=0}^N \frac{c_i  - b_i}{max(b_i , c_i)}
$$

$\bullet $ The \textit{Utility} evaluates the learning improvement when we learn a unique policy from all the state-action pairs w.r.t. learning $k$ policies from the identified clusters. Thus, an improvement in \textit{Utility} indicates that each cluster contains homogeneous state-action pairs in terms of strategy, as a model trained on them can predict the next action more easily. A higher utility indicates a better clustering. In particular, to evaluate the \textit{Utility} we compare the performance of a NN trained over all the state-action pairs w.r.t. to k NNs trained on the identified clusters. Let $\epsilon_i = ||\pi(s_i) - a_i||^2$ be the error for an observation $i$ when the action $a_i$ is predicted by a learned policy $\pi$. We denote with $\hat{\epsilon_i}$ the error when $i$ is predicted by a unique policy $\hat{\pi}$ trained on all the state-action pairs, while we denote with $\bar{\epsilon_i}$ the error when $i$ is predicted by a policy $\bar{\pi}$ trained on the cluster where $i$ belongs. Thus, the \textit{Utility} $\mathcal{U}$ can be defined as :
$$
\mathcal{U} = \frac{1}{N} \sum_{i=0}^N \frac{ \hat{\epsilon_i} - \bar{\epsilon_i} }{ \hat{\epsilon_i} }
$$

\section{Additional Experiments}
Here we present additional experiments that do not fit into the main body of the paper.

\subsection{Unknown number of clusters $k$}\label{app:k_unkwnn}
In this section, we evaluate the ability of {\algoname} to group the anonymous observations when the number of clusters $k$ is unknown. In such a case we adopt the classic Elbow method to optimize $k$ according to the \textit{distortion} of the clusters in the SHAP values space. The distortion measures the mean distance of each point to its assigned cluster.  

\begin{figure}[h]
\centering
\includegraphics[width=0.7\linewidth]{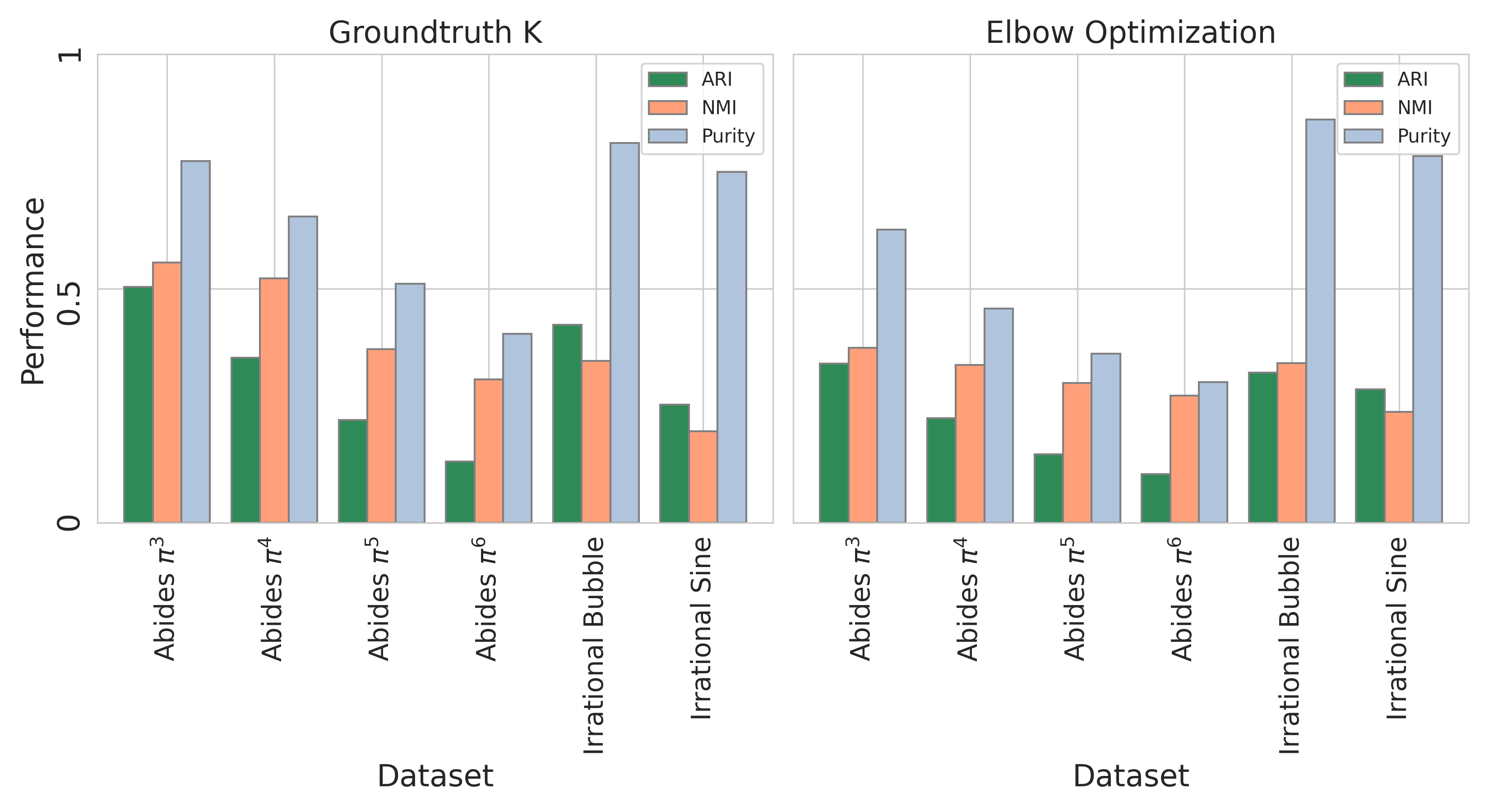}
\caption{K-SHAP - Elbow Optimization}\label{fig:elbow}
\end{figure}

Figure \ref{fig:elbow} shows the performance of {\algoname} when the ground truth number of clusters $k$ is given (left picture) and when $k$ is unknown (right picture). We consider only synthetic data as we do not know the ground truth $k$ for real market data. The picture shows that even if the performance is slightly inferior, {\algoname} maintains a similar trend and satisfactory performance when $k$ is unknown.

\newpage

\subsection{Contributions of the world-policy model}\label{app:wpolicies}
We now evaluate {\algoname} using three different models for the world-policy: 

$\bullet$ Random Forest - we use the implementation provided by scikit-learn~\cite{scikit-learn}, where we fix the number of trees to 100 and we use mean squared error as objective. 

$\bullet$ XGBoost~\cite{chen2016xgboost} - we use the implementation provided at \url{https://github.com/dmlc/xgboost}. We fix the max depth to 100, and we use mean squared error as objective.

$\bullet$ Neural Network - we implement a feedforward NN with 2-hidden layers using PyTorch. The architecture details are provided in Appendix \ref{app:nn_architecture}. 

For each world-policy we also consider a variant that uses a 2d embedding $\mathcal{Z}$ of the SHAP values, obtained through UMAP.

Table \ref{tab:wa_syn} and Table \ref{tab:wa_real} show the results for the different world-policies, for both synthetic and real data. The results on synthetic data confirm that a Random Forest world-policy better suits the anonymous market observations and provides better results.  However, also XGBoost achieves comparable performance, while the NN is not able to fully capture the structure of the data, achieving lower performance. The results are similar on the real historical data, even if XGBoost and NN outperform Random Forest for some stocks. 
In general, the higher the model accuracy in predicting the next action, given the input state, the better the SHAP values and related clustering.

\begin{table}[h]
\begin{minipage}{0.45\columnwidth}
\centering
   \resizebox{\linewidth}{!}{
   \begin{tabular}{lllll}
\toprule
Dataset & World-Policy &         ARI &        NMI &     Purity \\
\midrule
\multirow{7}{*}{\begin{tabular}[c]{@{}l@{}}
\textbf{Abides $\boldsymbol{\pi}^3$}\\ \cite{byrd2019abides} \end{tabular}}  & Random Forest &  \textbf{0.50±0.10} &  \textbf{0.56±0.05} &  \textbf{0.77±0.03} \\
                & Random Forest ($\mathcal{Z}$) &  0.23±0.00 &  0.25±0.00 &  0.60±0.00\\
                & NN &   0.03±0.02 &  0.05±0.01 &  0.42±0.02 \\
                & NN ($\mathcal{Z}$) &   0.20±0.01 &  0.23±0.01 &  0.58±0.01 \\
                & XGBoost &   0.48±0.01 &  0.55±0.01 &  0.76±0.00 \\
                & XGBoost ($\mathcal{Z}$) &   0.34±0.06 &  0.41±0.07 &  0.69±0.06 \\ \hline
\multirow{7}{*}{\begin{tabular}[c]{@{}l@{}}
\textbf{Abides $\boldsymbol{\pi}^4$}\\ \cite{byrd2019abides} \end{tabular}} & Random Forest &  \textbf{0.35±0.07} &  \textbf{0.52±0.02} &  \textbf{0.65±0.11} \\
                & Random Forest ($\mathcal{Z}$) &  0.21±0.08 &  0.32±0.03 &  0.55±0.03 \\
                & NN &   0.04±0.01 &  0.06±0.01 &  0.35±0.01 \\
                & NN ($\mathcal{Z}$) &   0.16±0.04 &  0.19±0.05 &  0.44±0.03 \\
                & XGBoost &   0.29±0.01 &  0.44±0.05 &  0.57±0.01 \\
                & XGBoost ($\mathcal{Z}$) &   0.24±0.01 &  0.35±0.01 &  0.56±0.00 \\ \hline
\multirow{7}{*}{\begin{tabular}[c]{@{}l@{}}
\textbf{Abides $\boldsymbol{\pi}^5$}\\ \cite{byrd2019abides} \end{tabular}}  & Random Forest &  \textbf{0.22±0.08} &  \textbf{0.37±0.06} &  \textbf{0.51±0.10} \\
                & Random Forest ($\mathcal{Z}$) &  0.20±0.03 &  0.33±0.01 &  0.50±0.06  \\
                & NN &   0.06±0.05 &  0.09±0.08 &  0.30±0.04 \\
                & NN ($\mathcal{Z}$) &   0.15±0.01 &  0.20±0.01 &  0.37±0.01 \\
                & XGBoost &   0.21±0.01 &  0.36±0.02 &  0.48±0.01 \\
                & XGBoost ($\mathcal{Z}$) &   0.20±0.01 &  0.32±0.02 &  0.48±0.01 \\ \hline
\multirow{7}{*}{\begin{tabular}[c]{@{}l@{}}
\textbf{Abides $\boldsymbol{\pi}^6$}\\ \cite{byrd2019abides} \end{tabular}}  & Random Forest &  \textbf{0.15±0.01} & \textbf{0.31±0.01} &  \textbf{0.42±0.01} \\
                & Random Forest ($\mathcal{Z}$) &    0.15±0.02 &  0.29±0.01 &  0.42±0.02  \\
                & NN &   0.12±0.00 &  0.20±0.01 &  0.31±0.00 \\
                & NN ($\mathcal{Z}$) &   0.13±0.01 &  0.17±0.01 &  0.32±0.01 \\
                & XGBoost &   0.15±0.01 &  0.31±0.02 &  0.41±0.01 \\
                & XGBoost ($\mathcal{Z}$) &   0.16±0.00 &  0.26±0.00 &  0.40±0.01 \\ \hline
\multirow{7}{*}{\begin{tabular}[c]{@{}l@{}}
\textbf{RL-Agents Bubble} \\ \cite{liu2022biased} \end{tabular}}  & Random Forest & 0.00±0.00 &  0.00±0.00 &  0.58±0.00  \\
                & Random Forest ($\mathcal{Z}$) & \textbf{0.42±0.20} &  \textbf{0.35±0.15} &  \textbf{0.81±0.10}\\
                & NN &  0.00±0.00 &  0.00±0.00 &  0.58±0.00 \\
                & NN ($\mathcal{Z}$) &   0.09±0.07 &  0.08±0.07 &  0.64±0.06 \\
                & XGBoost &   0.00±0.00 &  0.00±0.00 &  0.58±0.00 \\
                & XGBoost ($\mathcal{Z}$) &   0.00±0.00 &  0.00±0.00 &  0.58±0.00 \\ \hline
\multirow{7}{*}{\begin{tabular}[c]{@{}l@{}}
\textbf{RL-Agents Sine} \\ \cite{liu2022biased} \end{tabular}}  & Random Forest &  0.02±0.00 &  0.01±0.00 &  0.56±0.00 \\
                & Random Forest ($\mathcal{Z}$) &  \textbf{0.25±0.07} &  \textbf{0.20±0.06} &  \textbf{0.75±0.04} \\
                & NN &    0.00±0.00 &   0.00±0.00 &   0.56±0.00 \\
                & NN ($\mathcal{Z}$) &    0.05±0.00 &   0.04±0.00 &   0.62±0.00 \\
                & XGBoost &   0.02±0.00 &  0.01±0.00 &  0.56±0.00 \\
                & XGBoost ($\mathcal{Z}$) &   0.01±0.00 &  0.00±0.00 &  0.56±0.00 \\
\bottomrule
\end{tabular}}
\caption{World Policies - Multi-Agent Synthetic Data}\label{tab:wa_syn}
\end{minipage}
\hfill 
\begin{minipage}{0.45\columnwidth}
\centering
   \resizebox{\linewidth}{!}{
\begin{tabular}{llll}
\toprule
Stock & World-Policy &     Utility &  Silhouette \\
\midrule
\textbf{ADAP} & Random Forest &   \textbf{0.33±0.00} &   0.12±0.00 \\
     \cite{totalview}                & Random Forest ($\mathcal{Z}$) &   0.21±0.04 &   0.00±0.01 \\
                   & NN &  0.01±0.00 &   \textbf{0.22±0.01} \\
                   & NN ($\mathcal{Z}$) &  0.00±0.00 &  0.01±0.01 \\
                   & XGBoost &   0.33±0.01 &   0.11±0.00 \\
                   & XGBoost ($\mathcal{Z}$) &   0.28±0.05 &   0.01±0.01 \\ \hline
\textbf{AINV} & Random Forest &   0.19±0.06 &   \textbf{0.11±0.06} \\
      \cite{totalview}               & Random Forest ($\mathcal{Z}$) &   0.10±0.05 &   0.01±0.01 \\\
                   & NN &   0.00±0.00 &   0.06±0.02 \\
                   & NN ($\mathcal{Z}$) &   0.00±0.00 &   0.03±0.05 \\
                   & XGBoost &   \textbf{0.28±0.00} &   0.09±0.00 \\
                   & XGBoost ($\mathcal{Z}$) &   0.21±0.02 &   0.01±0.01 \\ \hline
\textbf{AVXL} & Random Forest &   \textbf{0.36±0.01} &   \textbf{0.09±0.00} \\
     \cite{totalview}                & Random Forest ($\mathcal{Z}$) &   0.28±0.06 &   0.04±0.05 \\
                   & NN &  0.00±0.00 &   0.05±0.00 \\
                   & NN ($\mathcal{Z}$) &   0.00±0.00 &   0.05±0.06 \\
                   & XGBoost &   \textbf{0.36±0.01} &   0.08±0.00 \\
                   & XGBoost ($\mathcal{Z}$) &   0.36±0.02 &   0.01±0.01 \\
\bottomrule
\end{tabular}}
\caption{World Policies - Historical Market Data}\label{tab:wa_real}
\end{minipage}
\end{table}

\newpage
\subsection{Historical market data at varying of number of clusters k}\label{app:vary_k}
In this section, we evaluate the \textit{Utility} and \textit{Silhouette} when we vary the number of clusters $k$ from 2 to 6. We use historical market data for AVXL stock. For clarity of the presentation, we remove from the picture the variants of K-Means and {\algoname} that use a dimensionality reduction technique, namely K-Means ($\mathcal{Z}$) and {\algoname} ($\mathcal{Z}$).

\begin{figure}[h]
\centering
\includegraphics[width=0.7\linewidth]{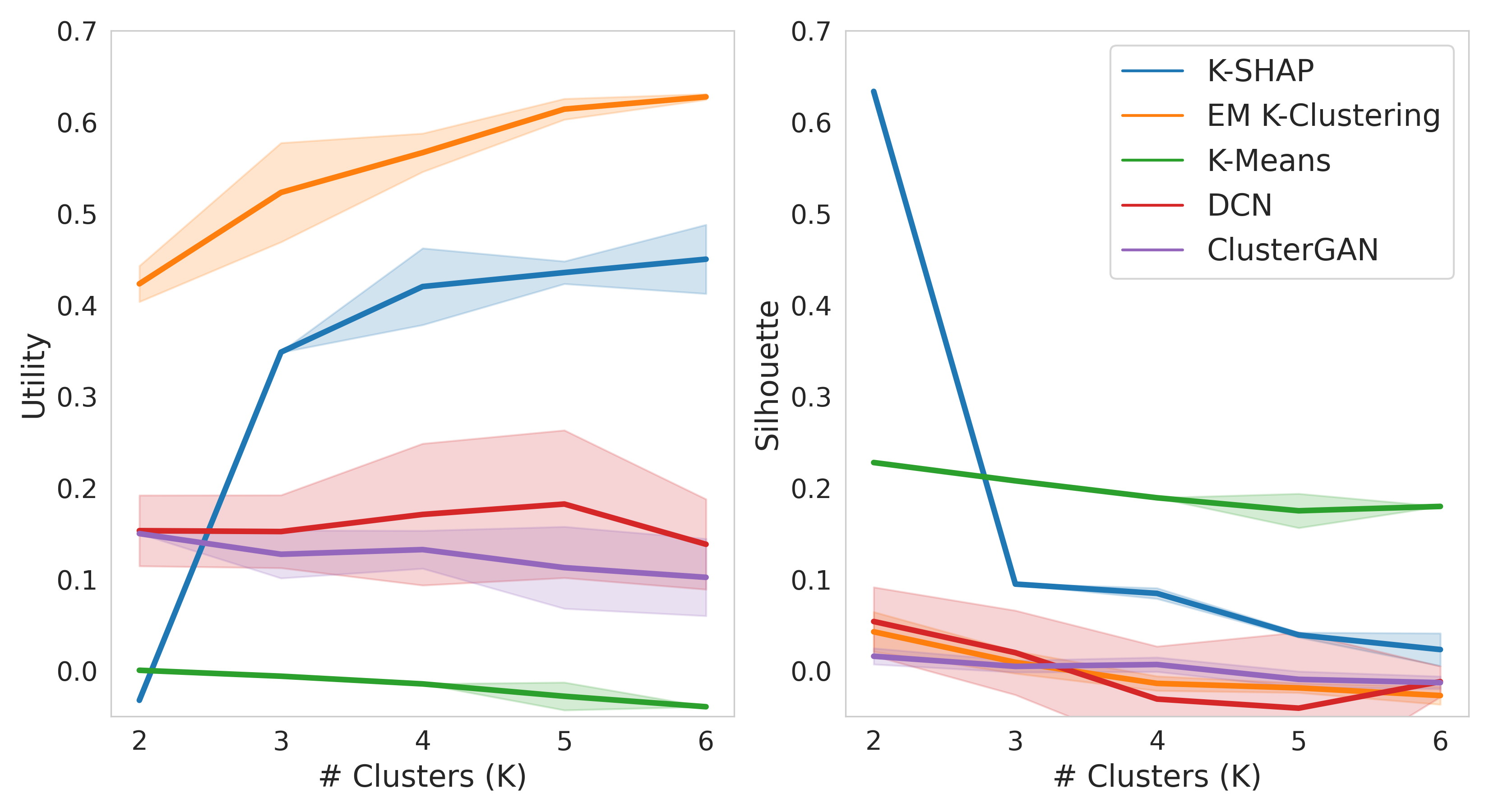}
\caption{Historical Market data AVXL at varying of nr. clusters k.}\label{fig:k_varying}
\end{figure}

Figure \ref{fig:k_varying} confirms the trend shown in the paper, and {\algoname} still represents a good trade-off between Utility and Silhouette. For $k=2$ K-SHAP groups the state-action pairs mostly into buy and sell orders. These two clusters improve the Silhouette, as the orders in each cluster are more homogeneous, but the Utility is close to 0. In fact, a cluster containing only buy (or sell) observations does not highlight any particular strategy. 

\newpage

\begin{table}[H]
\begin{minipage}{0.45\columnwidth}
\centering
   \resizebox{\linewidth}{!}{
\begin{tabular}{lrrrl}
\toprule
 Feature & mean & min & max & Type \\
\midrule 
Spread & 2.346 & 1.000 & 67.000 & Int \\
Vol. Imbalance Lev.1 & 0.500 & 0.001 & 0.999 & Float \\
Vol. Imbalance Lev.2 & 0.507 & 0.002 & 0.998 & Float \\
Vol. Imbalance Lev.5 & 0.513 & 0.007 & 0.997 & Float \\
Exec. Vol. Imbalance 1min & 0.462 & 0.000 & 1.000 & Float \\
Exec. Vol. Imbalance 5min & 0.463 & 0.187 & 0.820 & Float \\
Exec. Vol. Imbalance 12min & 0.463 & 0.255 & 0.765 & Float \\
Exec. Vol. Imbalance 26min & 0.463 & 0.289 & 0.725 & Float \\
Price return 1min & -0.000 & -0.002 & 0.002 & Float \\
Price return 5min & -0.000 & -0.003 & 0.004 & Float \\
Price return 12min & -0.000 & -0.003 & 0.006 & Float \\
Price return 26min & -0.000 & -0.005 & 0.007 & Float \\
Price MA 12s & 99.966 & 99.547 & 100.821 & Float \\
Price MA 26s & 99.966 & 99.532 & 100.822 & Float \\
Price MA 60s & 99.966 & 99.562 & 100.821 & Float \\
Price MA 5min & 99.966 & 99.586 & 100.814 & Float \\
Price MA 12min & 99.966 & 99.599 & 100.804 & Float \\
Price MA 26min & 99.967 & 99.644 & 100.793 & Float \\
Price MA 48min & 99.970 & 99.759 & 100.754 & Float \\
Price MA 1h36min & 99.976 & 99.832 & 100.567 & Float \\
Spread MA 12s & 1.599 & 1.000 & 49.083 & Float \\
Spread MA 26s & 1.594 & 1.000 & 57.192 & Float \\
Spread MA 60s & 1.595 & 1.000 & 39.950 & Float \\
Spread MA 5min & 1.601 & 1.000 & 21.200 & Float \\
Spread MA 12min & 1.606 & 1.000 & 13.250 & Float \\
Spread MA 26min & 1.611 & 1.000 & 7.885 & Float \\
Price MA Diff 12s-26s & 0.501 & 0.000 & 1.000 & Float \\
Price MA Diff 12min-26min & 0.392 & 0.000 & 1.000 & Float \\
Price MA Diff 48min-1h36min & 0.395 & 0.000 & 1.000 & Float \\ \hline
Order Size & 41.634 & 1.000 & 100.000 & Int \\
Order depth & 9.217 & -23.000 & 91.000 & Int \\
Order Direction & 0.495 & 0.000 & 1.000 & Bin \\
\bottomrule
\end{tabular}}
\caption{Summary and Description of Abides $\pi^3$}\label{tab:abides3}
\end{minipage}
\hfill
\begin{minipage}{0.45\columnwidth}    \centering
\resizebox{\linewidth}{!}{
 \centering
\begin{tabular}{lrrrl}
\toprule
 Feature & mean & min & max & Type \\
\midrule 
Spread & 2.647 & 1.000 & 67.000 & Int \\
Vol. Imbalance Lev.1 & 0.503 & 0.001 & 0.999 & Float \\
Vol. Imbalance Lev.2 & 0.511 & 0.002 & 0.998 & Float \\
Vol. Imbalance Lev.5 & 0.517 & 0.007 & 0.997 & Float \\
Exec. Vol. Imbalance 1min & 0.464 & 0.000 & 1.000 & Float \\
Exec. Vol. Imbalance 5min & 0.464 & 0.187 & 0.820 & Float \\
Exec. Vol. Imbalance 12min & 0.464 & 0.255 & 0.765 & Float \\
Exec. Vol. Imbalance 26min & 0.463 & 0.289 & 0.725 & Float \\
Price return 1min & -0.000 & -0.003 & 0.002 & Float \\
Price return 5min & 0.000 & -0.004 & 0.004 & Float \\
Price return 12min & 0.000 & -0.004 & 0.006 & Float \\
Price return 26min & 0.000 & -0.005 & 0.007 & Float \\
Price MA 12s & 99.970 & 99.534 & 100.821 & Float \\
Price MA 26s & 99.970 & 99.527 & 100.822 & Float \\
Price MA 60s & 99.970 & 99.562 & 100.821 & Float \\
Price MA 5min & 99.970 & 99.586 & 100.814 & Float \\
Price MA 12min & 99.970 & 99.599 & 100.804 & Float \\
Price MA 26min & 99.969 & 99.644 & 100.793 & Float \\
Price MA 48min & 99.971 & 99.759 & 100.754 & Float \\
Price MA 1h36min & 99.978 & 99.832 & 100.567 & Float \\
Spread MA 12s & 1.638 & 1.000 & 49.083 & Float \\
Spread MA 26s & 1.627 & 1.000 & 57.192 & Float \\
Spread MA 60s & 1.626 & 1.000 & 39.950 & Float \\
Spread MA 5min & 1.632 & 1.000 & 21.200 & Float \\
Spread MA 12min & 1.641 & 1.000 & 13.250 & Float \\
Spread MA 26min & 1.661 & 1.000 & 7.885 & Float \\
Price MA Diff 12s-26s & 0.512 & 0.000 & 1.000 & Float \\
Price MA Diff 12min-26min & 0.406 & 0.000 & 1.000 & Float \\
Price MA Diff 48min-1h36min & 0.395 & 0.000 & 1.000 & Float \\ \hline
Order Size & 43.883 & 1.000 & 100.000 & Int \\
Order depth & 6.441 & -23.000 & 91.000 & Int \\
Order Direction & 0.502 & 0.000 & 1.000 & Bin \\
\bottomrule
\end{tabular}}
\caption{Summary and Description of Abides $\pi^4$}\label{tab:abides4}
\end{minipage}
\vfill 
\begin{minipage}{0.45\columnwidth}
\centering
   \resizebox{\linewidth}{!}{
\begin{tabular}{lrrrl}
\toprule
 Feature & mean & min & max & Type \\
\midrule 

Spread & 2.779 & 1.000 & 67.000 & Int \\
Vol. Imbalance Lev.1 & 0.507 & 0.001 & 0.999 & Float \\
Vol. Imbalance Lev.2 & 0.509 & 0.004 & 0.998 & Float \\
Vol. Imbalance Lev.5 & 0.515 & 0.005 & 0.997 & Float \\
Exec. Vol. Imbalance 1min & 0.464 & 0.002 & 0.988 & Float \\
Exec. Vol. Imbalance 5min & 0.462 & 0.187 & 0.819 & Float \\
Exec. Vol. Imbalance 12min & 0.463 & 0.255 & 0.765 & Float \\
Exec. Vol. Imbalance 26min & 0.461 & 0.290 & 0.725 & Float \\
Price return 1min & -0.000 & -0.002 & 0.002 & Float \\
Price return 5min & -0.000 & -0.003 & 0.004 & Float \\
Price return 12min & -0.000 & -0.003 & 0.006 & Float \\
Price return 26min & -0.000 & -0.005 & 0.007 & Float \\
Price MA 12s & 99.964 & 99.564 & 100.821 & Float \\
Price MA 26s & 99.964 & 99.547 & 100.822 & Float \\
Price MA 60s & 99.964 & 99.577 & 100.821 & Float \\
Price MA 5min & 99.964 & 99.586 & 100.813 & Float \\
Price MA 12min & 99.964 & 99.599 & 100.804 & Float \\
Price MA 26min & 99.965 & 99.644 & 100.793 & Float \\
Price MA 48min & 99.968 & 99.759 & 100.754 & Float \\
Price MA 1h36min & 99.975 & 99.832 & 100.567 & Float \\
Spread MA 12s & 1.617 & 1.000 & 46.250 & Float \\
Spread MA 26s & 1.593 & 1.000 & 57.192 & Float \\
Spread MA 60s & 1.590 & 1.000 & 33.100 & Float \\
Spread MA 5min & 1.612 & 1.000 & 21.200 & Float \\
Spread MA 12min & 1.614 & 1.000 & 13.250 & Float \\
Spread MA 26min & 1.604 & 1.000 & 7.885 & Float \\
Price MA Diff 12s-26s & 0.506 & 0.000 & 1.000 & Float \\
Price MA Diff 12min-26min & 0.397 & 0.000 & 1.000 & Float \\
Price MA Diff 48min-1h36min & 0.374 & 0.000 & 1.000 & Float \\ \hline
Order Size & 42.638 & 1.000 & 100.000 & Int \\
Order depth & 4.688 & -23.000 & 91.000 & Int \\
Order Direction & 0.486 & 0.000 & 1.000 & Bin \\
\bottomrule
\end{tabular}}
\caption{Summary and Description of Abides $\pi^5$}\label{tab:abides5}
\end{minipage}
\hfill
\begin{minipage}{0.45\columnwidth}    \centering
\resizebox{\linewidth}{!}{
 \centering
\begin{tabular}{lrrrl}
\toprule
 Feature & mean & min & max & Type \\
\midrule 

Spread & 2.904 & 1.000 & 67.000 & Int \\
Vol. Imbalance Lev.1 & 0.507 & 0.001 & 0.999 & Float \\
Vol. Imbalance Lev.2 & 0.511 & 0.004 & 0.998 & Float \\
Vol. Imbalance Lev.5 & 0.516 & 0.005 & 0.997 & Float \\
Exec. Vol. Imbalance 1min & 0.465 & 0.002 & 1.000 & Float \\
Exec. Vol. Imbalance 5min & 0.463 & 0.187 & 0.819 & Float \\
Exec. Vol. Imbalance 12min & 0.463 & 0.255 & 0.765 & Float \\
Exec. Vol. Imbalance 26min & 0.462 & 0.290 & 0.725 & Float \\
Price return 1min & -0.000 & -0.003 & 0.002 & Float \\
Price return 5min & 0.000 & -0.004 & 0.004 & Float \\
Price return 12min & 0.000 & -0.004 & 0.006 & Float \\
Price return 26min & -0.000 & -0.005 & 0.007 & Float \\
Price MA 12s & 99.967 & 99.534 & 100.821 & Float \\
Price MA 26s & 99.967 & 99.527 & 100.822 & Float \\
Price MA 60s & 99.967 & 99.576 & 100.821 & Float \\
Price MA 5min & 99.968 & 99.586 & 100.813 & Float \\
Price MA 12min & 99.967 & 99.599 & 100.804 & Float \\
Price MA 26min & 99.967 & 99.644 & 100.793 & Float \\
Price MA 48min & 99.969 & 99.759 & 100.754 & Float \\
Price MA 1h36min & 99.976 & 99.832 & 100.567 & Float \\
Spread MA 12s & 1.635 & 1.000 & 46.250 & Float \\
Spread MA 26s & 1.614 & 1.000 & 57.192 & Float \\
Spread MA 60s & 1.608 & 1.000 & 33.100 & Float \\
Spread MA 5min & 1.634 & 1.000 & 21.200 & Float \\
Spread MA 12min & 1.644 & 1.000 & 13.250 & Float \\
Spread MA 26min & 1.643 & 1.000 & 7.885 & Float \\
Price MA Diff 12s-26s & 0.514 & 0.000 & 1.000 & Float \\
Price MA Diff 12min-26min & 0.408 & 0.000 & 1.000 & Float \\
Price MA Diff 48min-1h36min & 0.380 & 0.000 & 1.000 & Float \\ \hline
Order Size & 44.081 & 1.000 & 100.000 & Int \\
Order depth & 3.617 & -23.000 & 91.000 & Int \\
Order Direction & 0.489 & 0.000 & 1.000 & Bin \\
\bottomrule
\end{tabular}}
\caption{Summary and Description of Abides $\pi^6$}\label{tab:abides6}
\end{minipage}
\end{table}

\begin{table}[H]
\begin{minipage}{0.45\columnwidth}
\centering
   \resizebox{\linewidth}{!}{
\begin{tabular}{lrrrl}
\toprule
 Feature & mean & min & max & Type \\
\midrule 
Spread & 2.719 & 1.000 & 12.000 & Int \\
Vol. Imbalance Lev.1 & 0.477 & 0.001 & 0.999 & Float \\
Vol. Imbalance Lev.2 & 0.488 & 0.008 & 0.997 & Float \\
Vol. Imbalance Lev.5 & 0.482 & 0.070 & 0.938 & Float \\
Exec. Vol. Imbalance 1min & 0.615 & 0.000 & 1.000 & Float \\
Exec. Vol. Imbalance 5min & 0.663 & 0.000 & 1.000 & Float \\
Exec. Vol. Imbalance 12min & 0.671 & 0.000 & 1.000 & Float \\
Exec. Vol. Imbalance 26min & 0.650 & 0.034 & 1.000 & Float \\
Price return 1min & 0.000 & -0.008 & 0.011 & Float \\
Price return 5min & 0.000 & -0.014 & 0.012 & Float \\
Price return 12min & 0.000 & -0.022 & 0.016 & Float \\
Price return 26min & 0.000 & -0.030 & 0.021 & Float \\
Price MA 12s & 11.258 & 10.416 & 12.131 & Float \\
Price MA 26s & 11.257 & 10.419 & 12.133 & Float \\
Price MA 60s & 11.257 & 10.421 & 12.129 & Float \\
Price MA 5min & 11.257 & 10.430 & 12.131 & Float \\
Price MA 12min & 11.256 & 10.452 & 12.198 & Float \\
Price MA 26min & 11.256 & 10.465 & 12.252 & Float \\
Price MA 48min & 11.261 & 10.482 & 12.375 & Float \\
Price MA 1h36min & 10.380 & 10.492 & 11.471 & Float \\
Spread MA 12s & 2.506 & 1.000 & 10.333 & Float \\
Spread MA 26s & 2.517 & 1.000 & 9.846 & Float \\
Spread MA 60s & 2.536 & 1.000 & 8.300 & Float \\
Spread MA 5min & 2.664 & 1.000 & 9.400 & Float \\
Spread MA 12min & 2.802 & 1.000 & 11.500 & Float \\
Spread MA 26min & 3.035 & 1.154 & 18.154 & Float \\
Price MA Diff 12s-26s & 0.582 & 0.000 & 1.000 & Float \\
Price MA Diff 12min-26min & 0.516 & 0.000 & 1.000 & Float \\
Price MA Diff 48min-1h36min & 0.452 & 0.000 & 1.000 & Float \\ \hline
Order Size & 105.494 & 1.000 & 300.000 & Int \\
Order depth & 0.655 & -4.000 & 74.000 & Int \\
Order Direction & 0.550 & 0.000 & 1.000 & Bin \\
\bottomrule
\end{tabular}}
\caption{Summary and Description of AINV stock data}\label{tab:ainv}
\end{minipage}
\hfill
\begin{minipage}{0.45\columnwidth}    \centering
\resizebox{\linewidth}{!}{
 \centering
\begin{tabular}{lrrrl}
\toprule
 Feature & mean & min & max & Type \\
\midrule 
Spread & 3.038 & 1.000 & 12.000 & Int \\
Vol. Imbalance Lev.1 & 0.479 & 0.000 & 1.000 & Float \\
Vol. Imbalance Lev.2 & 0.478 & 0.001 & 0.999 & Float \\
Vol. Imbalance Lev.5 & 0.479 & 0.048 & 0.974 & Float \\
Exec. Vol. Imbalance 1min & 0.536 & 0.000 & 1.000 & Float \\
Exec. Vol. Imbalance 5min & 0.574 & 0.000 & 1.000 & Float \\
Exec. Vol. Imbalance 12min & 0.584 & 0.000 & 1.000 & Float \\
Exec. Vol. Imbalance 26min & 0.569 & 0.010 & 0.972 & Float \\
Price return 1min & -0.000 & -0.016 & 0.011 & Float \\
Price return 5min & -0.000 & -0.022 & 0.016 & Float \\
Price return 12min & -0.001 & -0.030 & 0.021 & Float \\
Price return 26min & -0.002 & -0.041 & 0.033 & Float \\
Price MA 12s & 5.560 & 5.132 & 6.020 & Float \\
Price MA 26s & 5.561 & 5.140 & 6.017 & Float \\
Price MA 60s & 5.561 & 5.140 & 6.015 & Float \\
Price MA 5min & 5.560 & 5.143 & 6.015 & Float \\
Price MA 12min & 5.561 & 5.145 & 6.010 & Float \\
Price MA 26min & 5.564 & 5.153 & 6.009 & Float \\
Price MA 48min & 5.569 & 5.160 & 5.997 & Float \\
Price MA 1h36min & 5.574 & 5.179 & 6.026 & Float \\
Spread MA 12s & 2.804 & 1.000 & 11.917 & Float \\
Spread MA 26s & 2.835 & 1.000 & 15.192 & Float \\
Spread MA 60s & 2.887 & 1.000 & 16.783 & Float \\
Spread MA 5min & 3.067 & 1.000 & 16.000 & Float \\
Spread MA 12min & 3.216 & 1.167 & 13.417 & Float \\
Spread MA 26min & 3.374 & 1.308 & 15.885 & Float \\
Price MA Diff 12s-26s & 0.547 & 0.000 & 1.000 & Float \\
Price MA Diff 12min-26min & 0.465 & 0.000 & 1.000 & Float \\
Price MA Diff 48min-1h36min & 0.517 & 0.000 & 1.000 & Float \\ \hline
Order Size & 149.949 & 1.000 & 600.000 & Int \\
Order depth & 1.599 & -4.000 & 94.000 & Int \\
Order Direction & 0.497 & 0.000 & 1.000 & Bin \\
\bottomrule
\end{tabular}}
\caption{Summary and Description of AVXL stock data}\label{tab:avxl}
\end{minipage}
\vfill 
\begin{minipage}{0.45\columnwidth}
\centering
   \resizebox{\linewidth}{!}{
\begin{tabular}{lrrrl}
\toprule
 Feature & mean & min & max & Type \\
\midrule 
Spread & 2.254 & 1.000 & 26.000 & Int \\
Vol. Imbalance Lev.1 & 0.459 & 0.000 & 1.000 & Float \\
Vol. Imbalance Lev.2 & 0.462 & 0.002 & 0.998 & Float \\
Vol. Imbalance Lev.5 & 0.475 & 0.068 & 0.967 & Float \\
Exec. Vol. Imbalance 1min & 0.568 & 0.000 & 1.000 & Float \\
Exec. Vol. Imbalance 5min & 0.582 & 0.000 & 1.000 & Float \\
Exec. Vol. Imbalance 12min & 0.574 & 0.000 & 1.000 & Float \\
Exec. Vol. Imbalance 26min & 0.565 & 0.004 & 1.000 & Float \\
Price return 1min & 0.001 & -0.027 & 0.025 & Float \\
Price return 5min & 0.002 & -0.038 & 0.053 & Float \\
Price return 12min & 0.004 & -0.054 & 0.056 & Float \\
Price return 26min & 0.007 & -0.069 & 0.077 & Float \\
Price MA 12s & 5.994 & 5.260 & 6.405 & Float \\
Price MA 26s & 5.993 & 5.260 & 6.404 & Float \\
Price MA 60s & 5.992 & 5.260 & 6.401 & Float \\
Price MA 5min & 5.990 & 5.281 & 6.393 & Float \\
Price MA 12min & 5.984 & 5.329 & 6.375 & Float \\
Price MA 26min & 5.974 & 5.373 & 6.391 & Float \\
Price MA 48min & 5.964 & 5.411 & 6.407 & Float \\
Price MA 1h36min & 5.953 & 5.448 & 6.433 & Float \\
Spread MA 12s & 2.193 & 1.000 & 23.667 & Float \\
Spread MA 26s & 2.199 & 1.000 & 26.192 & Float \\
Spread MA 60s & 2.238 & 1.000 & 28.350 & Float \\
Spread MA 5min & 2.397 & 1.000 & 25.200 & Float \\
Spread MA 12min & 2.688 & 1.000 & 27.333 & Float \\
Spread MA 26min & 3.098 & 1.000 & 18.077 & Float \\
Price MA Diff 12s-26s & 0.609 & 0.000 & 1.000 & Float \\
Price MA Diff 12min-26min & 0.605 & 0.000 & 1.000 & Float \\
Price MA Diff 48min-1h36min & 0.560 & 0.000 & 1.000 & Float \\ \hline
Order Size & 156.239 & 1.000 & 600.000 & Int \\
Order depth & 1.069 & -4.000 & 50.000 & Int \\
Order Direction & 0.546 & 0.000 & 1.000 & Bin \\
\bottomrule
\end{tabular}}
\caption{Summary and Description of ADAP stock data}\label{tab:adap}
\end{minipage}
\end{table}

\begin{table}[H]
\begin{minipage}{0.45\columnwidth}
\centering
   \resizebox{\linewidth}{!}{
\begin{tabular}{lrrrl}
\toprule
 Feature & mean & min & max & Type \\
\midrule 
Price Best Ask & 99.893 & 98.858 & 100.352 & Float \\
Price Best Bid & 99.892 & 98.857 & 100.350 & Float \\
Vol Best Ask & 51.604 & 0.000 & 1443.000 & Int \\
Vol Best Bid & 43.314 & 0.000 & 1000.000 & Int \\
Exec. Vol Ask 1min & 3.504 & 0.000 & 400.000 & Int \\
Exec. Vol Bid 1min & 3.800 & 0.000 & 298.000 & Int \\
Book depth Ask & 14.420 & 0.000 & 1378.000 & Int \\
Book depth Bid & 15.873 & 0.000 & 1342.000 & Int \\
Last Exec Price & 99.884 & 98.857 & 100.302 & Float \\
Volatility 30 min & 72.411 & 0.000 & 608.599 & Float \\
Spread & 1.013 & 0.000 & 2.000 & Int \\
Price & 99.892 & 98.857 & 100.350 & Float \\
Norm. Price MA 1min & 1.000 & 0.992 & 1.011 & Float \\
Norm. Price MA 2min & 1.000 & 0.992 & 1.012 & Float \\
Norm. Price MA 3min & 1.000 & 0.988 & 1.012 & Float \\
Norm. Price MA 5min & 1.000 & 0.987 & 1.012 & Float \\
Norm. Price MA 6min & 1.000 & 0.987 & 1.012 & Float \\
Norm. Price MA 10min & 1.000 & 0.987 & 1.013 & Float \\
Norm. Price MA 15min & 1.000 & 0.988 & 1.013 & Float \\
Norm. Price MA 30min & 1.000 & 0.987 & 1.013 & Float \\ \hline
Action & 0.319 & -4.000 & 4.000 & Int \\
\bottomrule
\end{tabular}}
\caption{Summary and Description of RL-Agents Bubble}\label{tab:irr_bubble}
\end{minipage}
\hfill
\begin{minipage}{0.45\columnwidth}    \centering
\resizebox{\linewidth}{!}{
 \centering
\begin{tabular}{lrrrl}
\toprule
 Feature & mean & min & max & Type \\
\midrule 
Price Best Ask & 100.088 & 98.932 & 101.217  & Float \\
Price Best Bid & 100.087 & 98.931 & 101.217  & Float \\
Vol Best Ask & 45.171 & 0.000 & 500.000 & Int \\
Vol Best Bid & 45.905 & 0.000 & 776.000 & Int \\
Exec. Vol Ask 1min & 3.199 & 0.000 & 113.000 & Int \\
Exec. Vol Bid 1min & 2.453 & 0.000 & 200.000 & Int \\
Book depth Ask & 116.077 & 0.000 & 3664.000 & Int \\
Book depth Bid & 71.810 & 0.000 & 1396.000 & Int \\
Last Exec Price & 100.057 & 98.931 & 101.024 & Float \\
Volatility 30 min & 276.131 & 0.000 & 744.593 & Float \\
Spread & 1.001 & 0.000 & 2.000 & Int \\
Price & 100.087 & 98.931 & 101.217 & Float \\
Norm. Price MA 1min & 1.000 & 0.990 & 2.002 & Float \\
Norm. Price MA 2min & 1.001 & 0.990 & 2.002 & Float \\
Norm. Price MA 3min & 1.001 & 0.989 & 2.002 & Float \\
Norm. Price MA 5min & 1.001 & 0.987 & 2.002 & Float \\
Norm. Price MA 6min & 1.001 & 0.987 & 2.002 & Float \\
Norm. Price MA 10min & 1.000 & 0.985 & 2.002 & Float \\
Norm. Price MA 15min & 1.001 & 0.986 & 2.002 & Float \\
Norm. Price MA 30min & 1.000 & 0.986 & 2.002 & Float \\ \hline
Action & 0.095 & -4.000 & 4.000 & Int \\
\bottomrule
\end{tabular}}
\caption{Summary and Description of RL-Agents Sine}\label{tab:irr_sine}
\end{minipage}

\end{table}

\end{document}